\documentclass[lettersize,journal]{IEEEtran}
\usepackage{amsmath,amsfonts}
\usepackage{algorithmic}
\usepackage{algorithm}
\usepackage{array}
\usepackage[caption=false,font=normalsize,labelfont=sf,textfont=sf]{subfig}
\usepackage{textcomp}
\usepackage{stfloats}
\usepackage{url}
\usepackage{verbatim}
\usepackage{graphicx}
\usepackage{cite}
\usepackage{hyperref}
\usepackage{amsmath,amssymb,amsfonts}
\usepackage{multirow}
\usepackage{makecell}
\usepackage[table]{xcolor}
\usepackage{threeparttable}
\usepackage{subfig}

\begin{document}

\title{SuperCL: Superpixel Guided Contrastive Learning for Medical Image Segmentation Pre-training}
\author{Shuang Zeng, Lei Zhu, Xinliang Zhang, Hangzhou He, Yanye Lu*
\thanks{This work was supported in part by the Natural Science Foundation of China (82371112, 623B2001, 62394311), 
in part by Beijing Municipal Natural Science Foundation (Z210008).}
\thanks{Shuang Zeng, Lei Zhu, Xinliang Zhang, Hangzhou He and Yanye Lu are with the Institute of Medical Technology, 
Peking University Health Science Center, Peking University, Beijing 100191, China, also with the 
Department of Biomedical Engineering, Peking University, Beijing 100871, China, also with the National Biomedical 
Imaging Center, Peking University, Beijing 100871,China, also with the Institute of Biomedical Engineering, 
Shenzhen Bay Laboratory, Shenzhen 5181071, China, and also with the Institute of Biomedical Engineering, 
Peking University Shenzhen Graduate School, Shenzhen 518055.}

\thanks{*The corresponding author: Yanye Lu, (e-mail: yanye.lu@pku.edu.cn)}
}

\markboth{IEEE Transactions on Neural Networks and Learning Systems}%
{Shell \MakeLowercase{\textit{et al.}}: A Sample Article Using IEEEtran.cls for IEEE Journals}


\maketitle

\begin{abstract}
Medical image segmentation is a critical yet challenging task, primarily due to the difficulty of obtaining extensive datasets of high-quality, expert-annotated images. Contrastive learning presents a potential but still problematic solution to this issue. Because most existing methods focus on extracting instance-level
or pixel-to-pixel representation, which ignores the characteristics between intra-image similar pixel groups. Moreover, when considering 
contrastive pairs generation, most SOTA methods mainly rely on manually setting thresholds, which requires a large number of gradient experiments and lacks efficiency and generalization.
To address these issues, we propose a novel contrastive learning approach named SuperCL for medical image segmentation pre-training.
Specifically, our SuperCL exploits the structural prior and pixel correlation of images by introducing two novel contrastive pairs generation strategies:
Intra-image Local Contrastive Pairs (ILCP) Generation and Inter-image Global Contrastive Pairs (IGCP) Generation. Considering superpixel cluster aligns well with the concept of contrastive pairs generation, we utilize the superpixel map to 
generate pseudo masks for both ILCP and IGCP to guide supervised contrastive learning.
Moreover, we also propose two modules named Average SuperPixel Feature Map Generation (ASP) and Connected Components Label Generation (CCL)
 to better exploit the prior structural information for IGCP. Finally, experiments on 8 medical image datasets indicate our SuperCL 
 outperforms existing 12 methods. {\itshape i.e.} Our SuperCL achieves a superior performance with more precise predictions from 
 visualization figures and 3.15\%, 5.44\%, 7.89\% DSC higher than the previous best results on MMWHS, CHAOS, Spleen with 10\% annotations. 
 Our code will be released after acceptance.
\end{abstract}


\section{Introduction}
\IEEEPARstart{M}{edical} image segmentation plays a crucial role in computer-aided diagnosis and offers numerous benefits for clinical use. And with the development of deep learning, fully supervised learning has produced remarkable outcomes in various image understanding tasks. However, when the annotations are insufficient due to time constraints, label costs and expertise limitations, traditional supervised learning methods usually experience substandard performance, particularly in medical image segmentation. 

Recently, self-supervised learning (SSL) presents a promising solution to this challenge: 
it offers a pre-training strategy that relies solely on unlabeled data to obtain a suitable initialization for training downstream tasks with limited annotations.
As a specific variant of SSL, contrastive learning (CL) \cite{chen2020simple,he2020momentum,GCL,zeng2021positional,DiRA}
has emerged as particularly effective in learning image-level features from large-scale unlabeled datasets, thus significantly reducing annotation costs.
The fundamental concept behind CL is attracting positive sample pairs while pushing away negative sample pairs by optimizing a model with InfoNCE loss. And CL aims to pre-train a powerful encoder or network so that it can be fine-tuned into an accurate model with limited annotated data in downstream tasks.  
CL methods in natural imaging domain have demonstrated remarkable success, including SimCLR \cite{chen2020simple}, 
MoCo \cite{he2020momentum}, SwAV \cite{caron2020unsupervised}, BYOL \cite{grill2020bootstrap}, SimSiam \cite{chen2021exploring} and WCL \cite{WCL}.

\begin{figure}[t]
  \centering
  \includegraphics[width=0.5\textwidth]{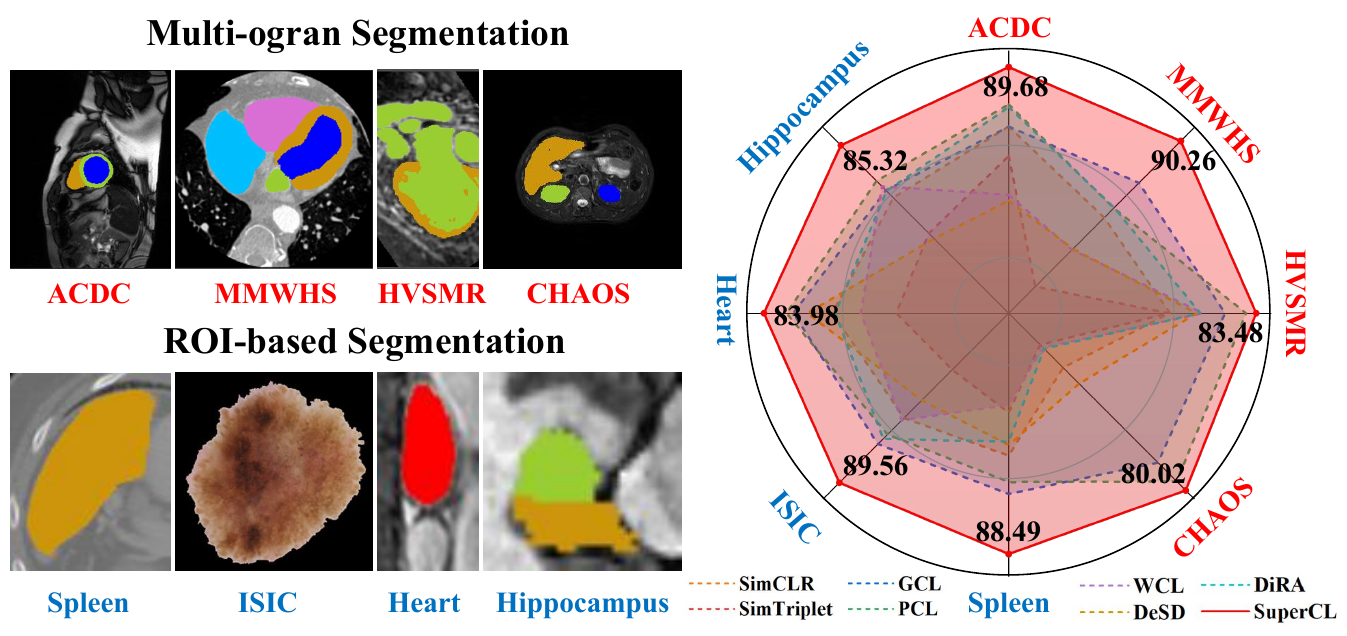}
  \caption{Our SuperCL (solid red lines) achieves SOTA segmentation performance (DSC) compared with other 7 CL baselines (dashed lines) across 4 multi-organ (red) and 4 ROI-based datasets (blue) with 25\% annotations.}
  \label{single-organ}
\end{figure}

In the realm of medical imaging, considerable attention \cite{GCL,zeng2021positional,DeSD,DiRA} has also been directed towards integrating unlabeled data to enhance network performance. 
In spite of CL framework design such as DeSD \cite{DeSD}, DiRA \cite{DiRA}, researchers also pay much attention to the elimination of false negative sample pairs including SimTriplet \cite{liu2021simtriplet}, 
GCL \cite{GCL} and PCL \cite{zeng2021positional}. 
Since similar anatomical structures often exist in all volumes of different patients in medical images such as CT and MRI, the corresponding 2D slices in different volumes often contain similar anatomical information which can be viewed as positive sample pairs. 
And with such a unique characteristic, SimTriplet \cite{liu2021simtriplet} constructs triplets from positive pairs, GCL \cite{GCL} proposes a {\itshape partition-based} contrasting strategy and PCL \cite{zeng2021positional} extends it into a {\itshape position-based} contrasting strategy.
However, despite their prevalence and success, existing CL methods still suffer from two major limitations. On the one hand, medical image segmentation aims to predict the class label for each pixel within one image, which focuses more on intra-image discrimination than inter-image discrimination. Based on this, it poses demands for more distinctive pixel-level (intra-image) representation learning, which will be more appropriate for medical image segmentation. However, most existing CL works focus on extracting instance-level projection \cite{chen2020simple,zeng2021positional,he2020momentum,DeSD} or 
just pixel-to-pixel projection \cite{GCL}, which ignores the characteristics between intra-image similar pixel groups. The former does not take pixel-level CL into account which is very important for dense prediction tasks like segmentation, while the latter only considers the simplest pixel-to-pixel CL pairs, hence introducing a large number of pixel-level (intra-image) false negative pairs. On the other hand, when considering instance-level (inter-image) contrastive pairs generation, the SOTA methods all rely on manually setting thresholds, 
such as position threshold of PCL \cite{zeng2021positional} and partition number of GCL \cite{GCL} to divide samples into positive and negative 
pairs which requires a large number of gradient experiments to verify the superiority of thresholds.

To address these limitations, we propose \textbf{SuperCL}, a novel CL approach which exploits the structural prior and pixel correlation of images by introducing two new contrastive
pairs generation strategies: Intra-image Local Contrastive Pairs Generation (ILCP) and Inter-image Global Contrastive Pairs Generation (IGCP).
Different from the previous CL works, there are mainly two characteristics of our proposed SuperCL: (1) Superpixel \cite{SLIC} can effectively group pixels with similar characteristics within the uniform regions of an image, which means pixels from the same cluster of superpixel map can be obviously and naturally viewed as positive pairs. Therefore, we can utilize the superpixel map to generate pseudo masks for both ILCP and IGCP to guide supervised CL. Benefited from this, we can not only incorporate pixel-level CL, but also minimize the number of false negative pairs. 
(2) We also propose two modules named Average SuperPixel Feature Map Generation (ASP) and Connected Components Label Generation (CCL) cooperating with the superpixel map to generate a more reliable weak label for IGCP to exploit the prior structural information better. 
Notably, superpixel map is only used for pre-training and not for fine-tuning. And both ILCP and IGCP do not require gradient experiments of manually setting thresholds. According to our experimental results, our proposed SuperCL outperforms other 12 methods across 8 medical image datasets partially shown in Fig. \ref{single-organ}. In summary, our main contributions are mainly threefold:
\begin{itemize}
\item A novel CL approach named SuperCL is proposed to exploit the structural prior and pixel correlation of images by introducing two new contrastive pairs generation strategies named ILCP and IGCP.

\item Superpixel map is utilized to generate pseudo masks for both ILCP and IGCP to guide supervised CL and two modules named ASP and CCL are proposed for better exploiting the prior structural information from the superpixel map.

\item State-of-the-art results have been achieved across 8 datasets from both CT and MRI for multi-organ segmentation and ROI-based segmentation compared with other 12 methods. Further we also conduct experiments to verify the generalization ability of our proposed SuperCL and ablation studies to validate the effectiveness of our proposed SuperCL.
\end{itemize}

\begin{figure*}[t]
  \centering
  \includegraphics[width=0.94\textwidth]{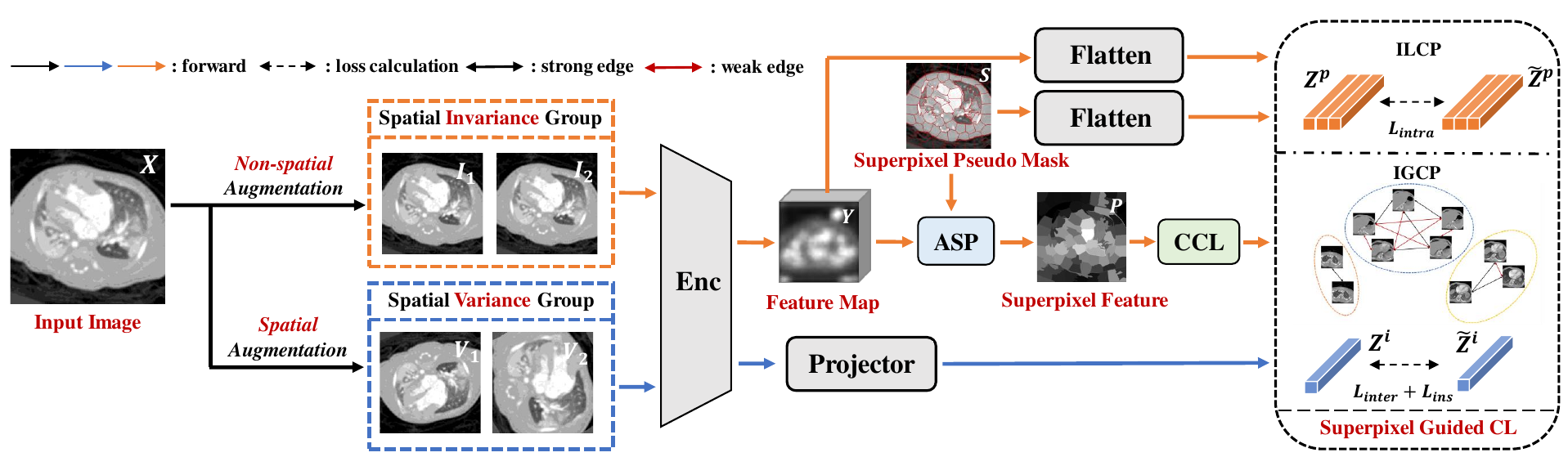}
  \caption{Overview of our proposed SuperCL. The image is input into two branches with two different augmentation settings: (1) The spatial invariance group is firstly fitted into the encoder to get the feature map. Then the feature map is flatten to get pixel-level projection. Finally guided with the superpixel pseudo mask (generated from SLIC and flatten), the pixel-level projection will be used for optimizing $\mathcal{L}_{intra}$ with ILCP (Sect. B). (2) The spatial variance group is propagated into the encoder along with a projector to get the instance-level projection. Then guided with a weak label, the instance-level projection will be used for optimizing $\mathcal{L}_{inter}$ with IGCP (Sect. C). Notably, the weak label is generated from the feature map and superpixel pseudo mask with two proposed modules: ASP and CCL (Sect. C). And $\mathcal{L}_{ins}$ functions as a baseline loss originated from WCL or PCL.}
  \label{SuperCL}
\end{figure*}

\section{Related Work}

\label{sec2}
\subsection{Medical Image Segmentation}

Over the past few years, deep learning, particularly convolutional neural networks (CNNs), has become an essential tool for medical image segmentation. Among those designs, FCN, UNet and DeepLab function as three milestones. They and their variants offer reliable baselines for segmentation tasks. Among these notable networks, 
U-Net \cite{ronneberger2015u} is renowned for its efficient encoder-decoder structures and skip connections, facilitating the accurate identification of anatomical structures.
Building upon U-Net's success, AttUNet \cite{oktay2018attention} integrates attention gates (AGs) into the skip connections to automatically filter out irrelevant areas in the input image and emphasize areas crucial for segmentation. 
Similarly, UCTransNet \cite{wang2022uctransnet} deals with semantic discrepancies in medical image segmentation by introducing a CTrans module that replaces the traditional skip connection.
And BCDUNet \cite{azad2019bi} incorporates a bidirectional convolutional LSTM module to 
nonlinearly integrate feature maps and employs dense convolution mechanisms to strengthen feature propagation and encourage feature reuse. 
ResUNet \cite{drozdzal2016importance} uses residual connections to facilitate information flow and alleviate vanishing gradient problems, leading to robust and stable segmentation outcomes.
Moreover, RollingUNet \cite{Liu_Zhu_Liu_Yu_Chen_Gao_2024} enhances feature integration by combining a multi-layer perceptron (MLP) with U-Net, effectively merging local features with long-range dependencies.
Additionally, the advent of Kolmogorov-Arnold Networks (KANs) also strengthens interpretability and efficiency through a series of learnable, non-linear activation functions. Therefore, UKAN \cite{li2024ukanmakesstrongbackbone} integrates KANs into the U-Net framework, enhancing its non-linear modeling capabilities and improving model interpretability.

\subsection{Superpixel Methods}

Traditional superpixel segmentation methods are broadly categorized into clustering-based and graph-based methods. The former utilizes classical clustering techniques like k-means to compute the connectivity between the anchor pixels and its neighbors, such as SLIC \cite{SLIC}, SNIC \cite{SNIC} and LSC \cite{LSC}. Specifically, SLIC \cite{SLIC} efficiently generates superpixel by limiting the search range of k-means. To further improve the efficiency, 
SNIC \cite{SNIC} is designed to update cluster center and arrange the label of pixels simultaneously through simple non-iterative clustering. 
Moreover, LSC \cite{LSC} is proposed to approximate the normalized cut energy by weighting k-means as linear spectral superpixel clustering. Different from clustering-based methods, graph-based methods firstly construct an undirected graph based on the feature of the input image and then generate superpixel by creating sub-graphs. Widely-used algorithms, such as Felzenszwalb and Huttenlocher (FH) \cite{FH} and entropy rate superpixel (ERS) \cite{ERS}, belong to this category. In FH \cite{FH}, each superpixel is represented by a minimum spanning tree and two superpixels are merged if the maximum weight of edges inside the trees is larger than the minimum weight of edges connecting them. While ERS \cite{ERS} maximizes the random walk entropy by continually adding edges into the graph model.

Recently, with the blossom of CNN, the CNN-based superpixel segmentation methods are emerging. They usually use CNN to extract features and then cluster the pixels based on these features. 
SEAL \cite{SEAL} firstly adopts CNN in superpixel segmentation by proposing a segmentation-aware loss. However, the overall architecture is not differentiable. Then SSN \cite{SSN} (superpixel sample network) develops the first differentiable deep network driven by the classic SLIC. But it still needs manual labels to supervise the network training and requires iteratively updating the predefined cluster centers to generate superpixels. SuperpixelFCN \cite{SuperpixelFCN} further simplifies the framework by assigning each pixel into the 9-neighbor grid, but it is still supervised by the segmentation labels. To address this issue, LNSNet \cite{LNSNet} attempts to learn the superpixels in an unsupervised manner and perform superpixel segmentation by a life-long clustering problem.    

\subsection{Contrastive Learning}
As a particular variant of SSL, CL has achieved remarkable success in extracting image-level features from large-scale unlabeled data which greatly reduces annotation costs, hence serving as a crucial role in video frame prediction, semantic segmentation, object detection and so on.
In practice, CL methods benefit from a large number of negative samples \cite{he2020momentum, chen2020simple}. 
MoCo \cite{he2020momentum} introduces a dynamic memory bank to store a queue of negative samples. 
SimCLR \cite{chen2020simple} directly uses a large batch size to make as many negative 
samples as possible coexist in the current batch. 
Beyond the aforementioned 
classic training frameworks, SwAV \cite{caron2020unsupervised} integrates online clustering into 
siamese network and proposes a novel `multi-crop' augmentation strategy that 
mixes the views of different resolutions. 
BYOL \cite{grill2020bootstrap} employs a slowly updated moving
average network, demonstrating that CL can be effective without 
any negative samples. By incorporating stop-gradient, SimSiam \cite{chen2021exploring} shows that 
even simple siamese networks are capable of learning meaningful representations without 
negative pairs, large batches or momentum encoders. 
Furthermore, WCL \cite{WCL} introduces a graph-based method to explore more positive sample pairs and generates a weak label for supervised contrastive learning.

In the medical imaging domain, substantial efforts of contrastive learning \cite{DeSD,DiRA,GCL,liu2021simtriplet,zeng2021positional} have been devoted to incorporating unlabeled data to improve network performance due to the limited data and annotations.
Researchers mainly focus on the design of siamese networks and construction of contrastive pairs to pre-train the model for several downstream tasks such as medical image classification, segmentation, registration and retrieval. 
Specifically, in the context of segmentation tasks,  
for the design of siamese networks, DeSD \cite{DeSD} restructures CL through a deep self-distillation approach, which enhances the representational capabilities of both the shallow and deep sub-encoders. This can tackle the challenge that SSL may suffer from the insufficient supervision at shallow layers.
DiRA \cite{DiRA} designs a novel self-supervised learning framework that integrates discriminative, restorative and adversarial learning in an unified manner to glean complementary visual information from unlabeled medical images. 
For the construction of contrastive pairs, 
SimTriplet \cite{liu2021simtriplet} introduces a {\itshape triplet-shape} CL framework. This framework aims to enhance both the intra-smple and inter-sample similarities by untilizing triplets derived from positive pairs, all without the need for negative samples.
GCL \cite{GCL} presents a 
{\itshape partition-based} contrasting strategy which leverages structural similarity across volumetric medical images. This strategy enables the division of unlabeled samples into positive and negative pairs and develops a localized contrastive loss function. 
Building on these concepts, PCL \cite{zeng2021positional} further proposes a {\itshape position-based} CL framework to generate contrastive data pairs by leveraging the position information in volumetric medical images.
This can help alleviate the issue where simple CL methods and GCL still introduce many false negative pairs, leading to degraded segmentation quality. This is especially problematic when different medical images contain similar structures.


\section{Methodology}\label{sec3}

\subsection{Overview}
The pipeline of our proposed SuperCL is illustrated in Fig. \ref{SuperCL}. Given an input 
image $\boldsymbol{X} \in \mathbb{R}^{C \times H \times W}$ ($H \times W$ is the resolution of the image, 
$C$ denotes the number of channels), two random augmentations $\boldsymbol{\tilde{X}}$ and $\boldsymbol{\hat{X}}$ of input image $\boldsymbol{X}$ will be obtained. 
In general setting, CL framework will pre-train an encoder $e(\cdot)$ along with a projection head $g(\cdot)$ and an instance-level contrastive loss $\mathcal{L}_{ins}$
among image-level sample pairs:
\begin{equation}
  \underset{\mathbf{\theta}}{\arg \min} \quad \mathcal{L}_{ins}  (g(e(\boldsymbol{\tilde{X}})), g(e(\boldsymbol{\hat{X}})))
\end{equation}
where $e(\cdot)$ and $g(\cdot)$ are two hypothetic functions that can be approximated by the network with learning parameter $\theta$.
Similarly, our proposed SuperCL is built upon the basic CL framework but with several adjustments. Firstly, two different settings for 
data augmentations will be applied to the input image $\boldsymbol{X}$: (1) One set is only non-spatial augmentations $T_{fix}$ 
which will be used for contrastive learning involving pixel-wise comparison $\mathcal{L}_{intra}$; (2) The other set is the combination of non-spatial and 
spatial augmentations $T_{var}$ which is suitable for instance-level contrastive learning $\mathcal{L}_{ins}$ and $\mathcal{L}_{inter}$.

\begin{equation}
  \begin{gathered}
    \{\boldsymbol{I_1}, \boldsymbol{I_2}\} = T_{fix}(\boldsymbol{X}) \quad
    \{\boldsymbol{V_1}, \boldsymbol{V_2}\} = T_{var}(\boldsymbol{X})
  \end{gathered}
\end{equation}

Secondly, the supervised signals consist of three losses: an instance-level loss $\mathcal{L}_{ins}$ like normal CL setting, 
an intra-image pixel-level loss $\mathcal{L}_{intra}$ (Section 2.2) and an inter-image 
instance-level loss $\mathcal{L}_{inter}$ (Section 2.3). Specifically, \textbf{spatial invariance group} $\{\boldsymbol{I_1}, \boldsymbol{I_2}\}$
will be propagated to the encoder $e(\cdot)$ to get feature map and then flatten to get pixel-level projection 
$\{{\boldsymbol{Z}}^{p}, \tilde{\boldsymbol{Z}}^{p} \} \in \mathbb{R}^{B\times C \times (h \times w)}$ 
(where $B$ denotes batch size, $C$ denotes channels, $h$ denotes height of the feature map, $w$ denotes width of the feature map). While \textbf{spatial variance group} $\{\boldsymbol{V_1}, \boldsymbol{V_2}\}$ will be propagated to the encoder $e(\cdot)$ and the projector
$g(\cdot)$ to get instance-level projection $\{{\boldsymbol{Z}}^{i}, \tilde{\boldsymbol{Z}}^{i} \} \in \mathbb{R}^{B\times C}$:

\begin{equation}
  \begin{gathered}
    {\boldsymbol{Z}}^{p} = \text{Flatten}(e(\boldsymbol{I_1})) \quad
    \tilde{\boldsymbol{Z}}^{p}=\text{Flatten}(e(\boldsymbol{I_2})) \\ 
    {\boldsymbol{Z}}^{i} = g(e(\boldsymbol{V_1})) \quad
    \tilde{\boldsymbol{Z}}^{i}=g(e(\boldsymbol{V_2}))
  \end{gathered}
  \end{equation}

\begin{figure}[t]
  \centering
  \includegraphics[width=0.5\textwidth]{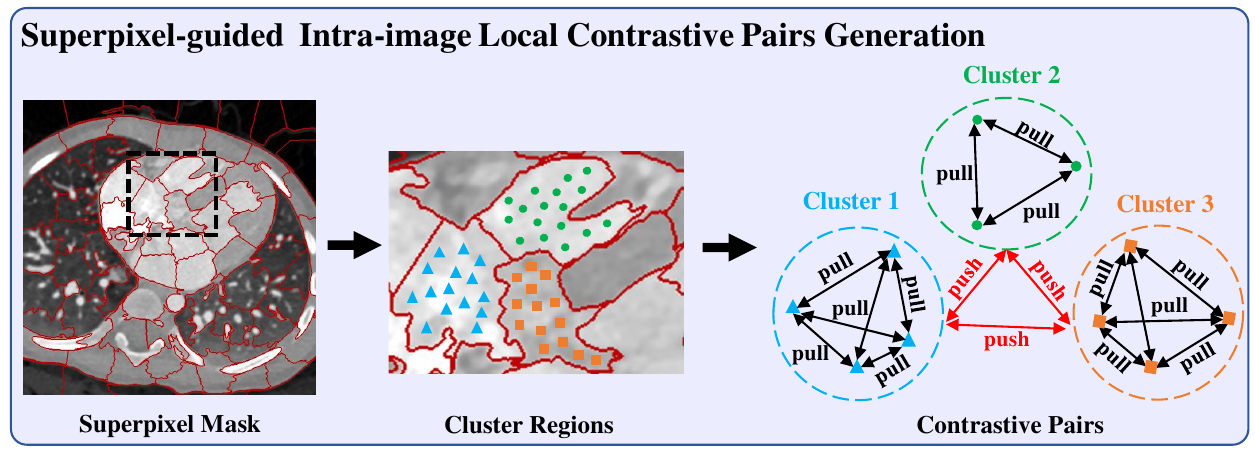}
  \caption{Superpixel-guided intra-image local contrastive pairs generation.}
  \label{ILCP}
\end{figure}

Finally, we can optimize the network with SuperCL with the above three contrastive losses considering both instance-level and pixel-level representations.
Specifically, the contrastive loss can be formulated as:
\begin{equation}
  \begin{gathered}
  \mathcal{L}_{total}=\lambda_1 \mathcal{L}_{ins}+\lambda_2 \mathcal{L}_{intra}+\lambda_{3} \mathcal{L}_{inter}  
  \end{gathered} 
  \end{equation}
where $\lambda_1, \lambda_2, \lambda_3$ are the weighting factors. $\mathcal{L}_{ins}$ can be viewed as a fundamental baseline loss and we propose
another two strategies: ILCP for $\mathcal{L}_{intra}$ and IGCP for $\mathcal{L}_{inter}$ which will be discussed in the following two sections.
\begin{figure*}[t]
  \centering
  \includegraphics[width=\textwidth]{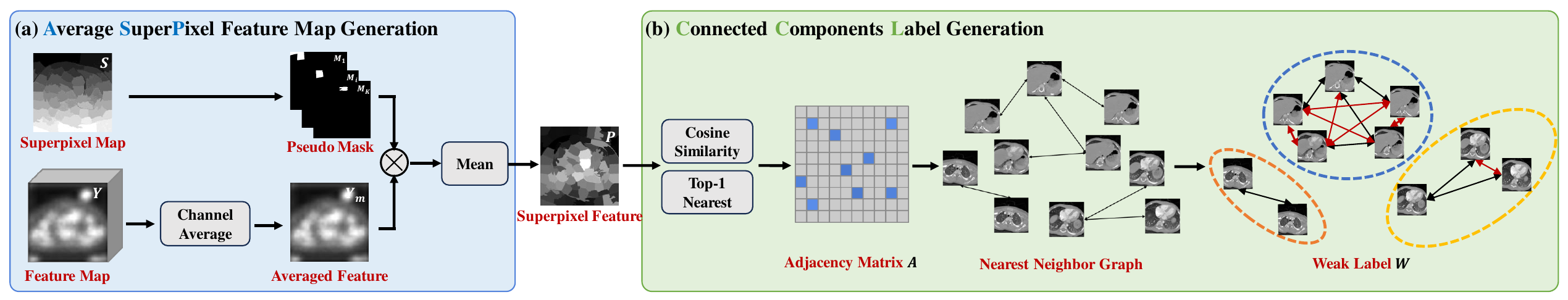}
  \caption{Illustration of our proposed ASP and CCL modules. (a) ASP aims at generating a more reliable representation for affinity matrix calculation. (b) CCL aims at transforming the representation into
  the weak label used for supervised CL.}
  \label{ASPCCL}
\end{figure*}

\subsection{Intra-image Local Contrastive Pairs Generation}

As shown in Fig.\ref{ILCP}, considering that superpixel can effectively group pixels with similar characteristics within the uniform regions of an 
image which means pixels from the same cluster of superpixel map can be obviously and naturally viewed as positive pairs, 
we utilize the superpixel of the image as a guide for Intra-image Local Contrastive Pairs (ILCP) generation.  
Since the pixel-level projection is spatially invariable, thus the flatten superpixel map corresponds to the pixel-level 
projection pixel-by-pixel if they are from the same input image. And the superpixel map can be used to guide ILCP generation as a pesudo mask for 
supervised CL:


\begin{equation}
\begin{aligned}
\Omega_{\textbf{ILCP}}:& \quad \{\boldsymbol{Z}_i^p, \tilde{\boldsymbol{Z}}_j^p\} \in \Omega_{\textbf{ILCP}}, \text { if } i, j \in \boldsymbol{S}_k, \ k \in[1, K] \\
\mathcal{L}_{intra} &= \sum_{i=1}^{2 h \times w} -\frac{1}{\left|\Omega_i^{+}\right|} \sum_{j \in \Omega_i^{+}} 
\log \frac{e^{s i m\left(Z_{i}^p, \tilde{Z}_{j}^p\right) / \tau}}{\sum_{k=1}^{2 h \times w} \mathbb{I}_{i \neq k} 
\cdot e^{s i m\left(Z_{i}^p, \tilde{Z}_{k}^p\right) / \tau}},\\
& \quad \quad \quad \quad \quad \quad \Omega^{+} = \Omega_{\textbf{ILCP}}
\end{aligned}
\end{equation}
where $\boldsymbol{S}_k$ denotes $k_{th}$ cluster of the superpixel map $\boldsymbol{S}$, $K$ denotes the total cluster number of 
superpixel.
$h\times w$ means the total number of pixels in the feature map. 
$\left|\Omega^{+}\right|$ is the set of indices of positive samples. $sim(\cdot, \cdot)$ is the cosine similarity function. $\tau$ is a temperature scaling parameter.
$\mathbb{I}$ is an indicator function.
According to the definition of ILCP, we can infer that if the corresponding pixel $i$ of pixel-level projection $\boldsymbol{Z}_i^p$ and pixel $j$ of pixel-level projection $\tilde{\boldsymbol{Z}}_j^p$
are from the same superpixel cluster $\boldsymbol{S}_k$, they will be viewed as positive pairs.

\subsection{Inter-image Global Contrastive Pairs Generation}

Inspired by WCL \cite{WCL}, which models each batch of instances as a nearest neighbor graph to generate a weak label for supervised CL, we also aim to leverage image superpixels to generate a more reliable weak label. This label will be better suited for medical image datasets by expanding instance-level (inter-image) global positive contrastive pairs for supervised CL.
Specifically, we propose ASP and CCL modules cooperating with the superpixel map 
for better exploiting the prior structural information for IGCP. 


\noindent \textbf{Average SuperPixel Feature Map Generation (ASP):} As illustrated in Fig. \ref{ASPCCL}(a), 
instead of measuring cosine similarity between the instance-level projection $\boldsymbol{Z}^{i}$ and $\tilde{\boldsymbol{Z}}^{i}$ 
like WCL, we propose ASP module to generate a more reliable representation for cosine similarity calculation. 
Specifically, we get the superpixel
map $\boldsymbol{S} \in \mathbb{R}^{B \times h \times w}$ and feature map $\boldsymbol{Y} \in \mathbb{R}^{B \times C \times h \times w}$
from spatial invariance group $\{\boldsymbol{I_1}, \boldsymbol{I_2}\}$. Then we obtain a binary pseudo mask 
$\boldsymbol{M}_c \in \mathbb{R}^{B \times h \times w}, c \in [1, K]$ for each cluster from the superpixel
map, perform element-wise multiplication with the feature map averaged across channels $\boldsymbol{Y}_m \in \mathbb{R}^{B \times h \times w}$ and compute
the averaged pixel value. 
Notably, the resulting avereged pixel value will serve as the pixel value in the region of our new Averaged SuperPixel Feature $\boldsymbol{P} \in \mathbb{R}^{B \times h \times w}$
which corresponds to the foreground region (where $\text{pixel value} = 1$) in the binary mask $\boldsymbol{M}_i$:
\begin{equation}
\begin{aligned}
&\boldsymbol{P}_c = \frac{1}{h \times w}\sum_{y=0}^{w-1}\sum_{x=0}^{h-1}[\boldsymbol{Y}_m(x,y) \cdot \boldsymbol{M}_{c}(x, y)], \ c \in [1, K] \\
&\boldsymbol{Y} = e(\boldsymbol{I}), \quad \boldsymbol{Y}_m = \frac{1}{C} \sum_{i=1}^C \boldsymbol{Y}_i, \quad 
\boldsymbol{S} = down(s(\boldsymbol{I})) \\
&\boldsymbol{M}_c(x, y)= 
\begin{cases}
1, \text{ if } \boldsymbol{S}(x,y)=c, \ x \in [0, h-1], y \in [0, w-1] \\
0, \text{ otherwise }
\end{cases}
\end{aligned}
\end{equation}
where $s(\cdot)$ denotes SLIC algorithm, $down(\cdot)$ denotes downsampling used for size alignment between the superpixel map $\boldsymbol{S}$ 
and the feature map $\boldsymbol{Y}_m$. $(x, y)$ denotes the pixel position. All these $\boldsymbol{P}_c$ form the final $\boldsymbol{P}$. According to ASP module, averaged superpixel feature $\boldsymbol{P}$ 
can retain the pixel-wise information of the original feature map $\boldsymbol{Y}$ which also relates to the instance-level projection 
( $\boldsymbol{I}_1 \rightarrow \boldsymbol{V}_1, \boldsymbol{I}_2 \rightarrow \boldsymbol{V}_2$). 
Moreover, it also implants the category feature of the
corresponding superpixel map $\boldsymbol{S}$.

\noindent \textbf{Connected Components Label Generation (CCL):} After we get the averaged superpixel feature $\boldsymbol{P}$,
we can utilize the CCL module to generate a more reliable weak label as shown in Fig. \ref{ASPCCL}(b). Specifically, similar to WCL, 
for each sample $\boldsymbol{P}_i \in \mathbb{R}^{h \times w}$ from $\boldsymbol{P}$ in the mini-batch (size is $B$), we find the closest sample $\boldsymbol{P}_j$ by computing the cosine similarity score $\boldsymbol{C}_{ij}$.
Then we can define an adjacency matrix $\boldsymbol{A}$ by Top-1 Nearest as follows:
\begin{equation}
\begin{aligned}
&\boldsymbol{C}_{ij} = sim(\boldsymbol{P}_i, \boldsymbol{P}_j), \quad j \neq i, \text{ and } j \in [1, B] \\
&\boldsymbol{A}(i, j) = \begin{cases}
1, & \text{ if } \boldsymbol{C}_{ij} = \max_{k} \, \boldsymbol{C}_{ik}, k \neq j \\
  & \text{ or } \boldsymbol{C}_{ji} = \max_{k} \, \boldsymbol{C}_{jk}, k \neq i \\
0, & \text{ otherwise }
\end{cases}
\end{aligned}
\label{adjacency}
\end{equation}

Basically, Eq. \ref{adjacency} will generate a sparse and symmetric 1-nearest neighbor graph where each vertex is linked with its closest sample.
Furthermore, for each sample, we will find all the reachable samples based on the 1-nearest neighbor graph by the Hoshen-Kopelman algorithm
to get the weak label $\boldsymbol{W}$ as follows:
  \begin{equation}
    \begin{gathered}
    \boldsymbol{W}(i, j)= \begin{cases}1, & \text { if } \text{find}(\boldsymbol{P}_i)= \text{find}(\boldsymbol{P}_j) \text { and } i \neq j
    \\ 0, & \text { otherwise }\end{cases}
    \end{gathered}
    \end{equation}
where $\text{find}(\boldsymbol{P}_i)= \text{find}(\boldsymbol{P}_j)$ means $\boldsymbol{P}_i$ and $\boldsymbol{P}_j$ belong to the same 
connected component of the undirected graph, and hence can be viewed as positive contrastive pairs for supervised CL. Finally, simlar to $\mathcal{L}_{ins}$, $\mathcal{L}_{inter}$ can be 
formulated as:
\begin{equation}
  \begin{aligned}
\mathcal{L}&=\sum_{l=1}^{2N} -\frac{1}{\left|\Omega_l^{+}\right|} \sum_{j \in \Omega_l^{+}} 
\log \frac{e^{s i m\left(Z_{l}^i, \tilde{Z}_{j}^i\right) / \tau}}{\sum_{k=1}^{2 N} \mathbb{I}_{l \neq k} 
\cdot e^{s i m\left(Z_{l}^{i}, \tilde{Z}_{k}^{i}\right) / \tau}}\\
\Omega^{+} &= \Omega_{\textbf{W}} (\textbf{IGCP}), \, \text{ if } \mathcal{L}=\mathcal{L}_{inter}, \  
\Omega^{+} = \Omega_{\textbf{PCL}}, \ \text{ if } \mathcal{L}=\mathcal{L}_{ins}
\end{aligned}
\end{equation}
where $N$ denotes the number of randomly sampled slices from the dataset, $\boldsymbol{Z}^i$ and $\tilde{\boldsymbol{Z}}^i$ are instance-level projection.


\begin{table*}[t]
\centering
\begin{threeparttable}
\caption{Quantitative results of our SuperCL against other SOTA methods on 4 multi-organ segmentation datasets.}
\label{mutiorgan}
\scriptsize
\setlength{\tabcolsep}{1pt}
\begin{tabular}{c|c|cccc|cccc|cccc|cccc}
\Xhline{1.0pt} 
 &  & \multicolumn{4}{c|}{ ACDC (100 patients) } & \multicolumn{4}{c|}{ MMWHS (20 patients) } & \multicolumn{4}{c|}{ HVSMR (10 patients) } & \multicolumn{4}{c}{ CHAOS (20 patients)} \\
 {Labeled}& {Methods} & DSC (\%) $\uparrow$ & JC (\%) $\uparrow$ & HD95 $\downarrow$ & ASD $\downarrow$ & DSC (\%) $\uparrow$ & JC (\%) $\uparrow$ & HD95 $\downarrow$ & ASD $\downarrow$ 
 & DSC (\%) $\uparrow$ & JC (\%) $\uparrow$ & HD95 $\downarrow$ & ASD $\downarrow$
 & DSC (\%) $\uparrow$ & JC (\%) $\uparrow$ & HD95 $\downarrow$ & ASD $\downarrow$
 \\
\Xhline{1.0pt} \multirow{12}*{10 \%}
& Scratch &$79.99$ &$67.39$ &$8.78$ &$3.82$ &$63.04$ &$48.89$ &$9.31$ &$3.39$ 
&$72.81$ &$59.02$ &$34.34$ &$10.18$ &$51.76$ &$37.51$ &$23.08$ &$9.90$ 
\\
& SimCLR \cite{chen2020simple} &$78.60$ &$65.47$ &$8.11$ &$3.36$ &$61.70$ &$46.59$ &$9.67$ &$3.53$ 
&$75.78$ &$62.34$ &$25.16$ &${\mathbf{6.22}}$ &$53.22$ &$38.11$ &$22.28$ &$9.04$
\\
& MoCo \cite{he2020momentum} &$81.34$ &$69.13$ &$8.12$ &$3.20$ &$57.54$ &$42.62$ &$11.33$ &$4.15$
&$76.20$ &$62.79$ &$27.45$ &$8.33$ &$53.14$ &$38.49$ &$21.45$ &$9.53$
\\
& BYOL \cite{grill2020bootstrap} &$77.55$ &$64.17$ &$9.05$ &$3.96$ &$63.75$ &$49.54$ &$9.34$ &$3.61$ 
&$73.57$ &$59.87$ &$29.79$ &$8.51$ &$50.45$ &$35.66$ &$24.20$ &$10.32$
\\
& SwAV \cite{caron2020unsupervised} &$82.11$ &$70.20$ &$\underline{6.35}$ &$\underline{2.46}$ &$64.27$ &$49.96$ &$10.31$ &$3.98$ 
&$\underline{76.28}$ &$\underline{62.91}$ &$24.09$ &$6.42$ &$49.87$ &$35.04$ &$19.75$ &$8.86$
\\
& SimSiam \cite{chen2021exploring} &$71.96$ &$57.01$ &$11.65$ &$4.94$ &$61.37$ &$46.51$ &$10.61$ &$3.89$
&$75.10$ &$61.47$ &$29.73$ &$8.72$ &$41.21$ &$28.87$ &$21.55$ &$10.20$
\\
& SimTriplet \cite{liu2021simtriplet} &$71.07$ &$56.07$ &$11.31$ &$5.20$ &$62.52$ &$47.81$ &$10.98$ &$4.10$ 
&$72.66$ &$58.48$ &$30.72$ &$8.65$ &$32.88$ &$23.14$ &$27.81$ &$12.90$
\\
& GCL \cite{GCL} &$84.86$ &$74.07$ &$7.07$ &$2.85$ &$\underline{71.41}$ &$\underline{58.58}$ &$\underline{7.27}$ &$\underline{2.55}$
&$75.70$ &$62.22$ &$28.15$ &$7.68$ &$58.78$ &$44.00$ &${\mathbf{15.28}}$ &$\underline{6.36}$
\\
& PCL \cite{zeng2021positional} &$\underline{84.91}$ &$\underline{74.16}$ &$6.54$ &$2.65$ &$65.98$ &$52.32$ &$8.07$ &$3.16$
&$76.12$ &$62.82$ &$24.50$ &$6.43$ &$\underline{63.61}$ &$\underline{48.23}$ &${16.15}$ &${6.79}$
\\
& WCL \cite{WCL} &$75.46$&$61.52$ &$11.05$&$4.69$&$60.82$ &$45.66$ &$9.82$ &$3.92$ 
&$75.05$ &$61.42$ &$26.09$ &$\underline{6.35}$ &$56.87$ &$41.39$ &$20.83$ &$9.31$
\\
& DeSD \cite{DeSD} &$76.02$ &$62.27$ &$10.25$ &$4.15$ &$57.03$ &$42.30$ &$10.89$ &$4.36$
&$74.57$ &$61.16$ &$\underline{24.00}$ &$6.40$ &$61.93$ &${46.68}$ &$18.91$ &$8.39$
\\
& DiRA \cite{DiRA} &$83.14$ &$71.61$ &$6.73$ &$2.60$ &$57.73$ &$42.95$ &$10.92$ &$3.76$
&$74.87$ &$61.35$ &$27.91$ &$8.01$ &$44.30$ &$31.24$ &$24.63$ &$11.93$
\\
& \cellcolor{cyan!20!green!20!}SuperCL (Ours) &\cellcolor{cyan!20!green!20!}$\mathbf{86.06}$ &\cellcolor{cyan!20!green!20!}$\mathbf{75.79}$ 
&\cellcolor{cyan!20!green!20!}$\mathbf{4.68}$ 
&\cellcolor{cyan!20!green!20!}$\mathbf{1.80}$ 
&\cellcolor{cyan!20!green!20!}$\mathbf{74.56}$ 
&\cellcolor{cyan!20!green!20!}$\mathbf{62.05}$ 
&\cellcolor{cyan!20!green!20!}$\mathbf{6.50}$ 
&\cellcolor{cyan!20!green!20!}$\mathbf{2.22}$ 
&\cellcolor{cyan!20!green!20!}$\mathbf{78.35}$ 
&\cellcolor{cyan!20!green!20!}$\mathbf{65.46}$ 
&\cellcolor{cyan!20!green!20!}$\mathbf{23.23}$ 
&\cellcolor{cyan!20!green!20!}$6.61$ 
&\cellcolor{cyan!20!green!20!}$\mathbf{69.05}$ 
&\cellcolor{cyan!20!green!20!}$\mathbf{54.09}$ 
&\cellcolor{cyan!20!green!20!}$\underline{15.43}$ 
&\cellcolor{cyan!20!green!20!}$\mathbf{6.32}$
\\
\Xhline{1.0pt} 
\multirow{12}*{25 \%}
& Scratch &$88.99$ &$80.34$ &$\underline{3.93}$ &$\underline{1.24}$ &$85.14$ &$74.87$ &$4.66$ &$1.45$
&$81.26$ &$69.25$ &$20.63$ &$5.85$ &$66.71$ &$52.74$ &$12.32$ &$5.39$
\\
& SimCLR \cite{chen2020simple} &$88.55$ &$79.71$ &$4.53$ &$1.47$ &$86.14$ &$76.03$ &$4.59$ &$1.36$
&$80.53$ &$68.50$ &$23.42$ &$6.75$ &$66.15$ &$52.21$ &$11.76$ &$4.75$
\\
& MoCo \cite{he2020momentum} &$88.38$ &$79.39$ &$5.01$ &$1.55$ &$82.89$ &$71.46$ &$5.40$ &$1.69$
&$81.26$ &$69.18$ &$20.41$ &$5.64$ &$65.33$ &$50.93$ &$15.12$ &$6.40$
\\
& BYOL \cite{grill2020bootstrap} &$\underline{89.27}$ &$\underline{80.78}$ &$4.62$ &$1.64$ &$87.63$ &$78.43$ &$3.83$ &$1.20$
&$81.33$ &$69.30$ &$19.95$ &$5.80$ &$62.49$ &$48.31$ &$14.13$ &$5.71$
\\
& SwAV \cite{caron2020unsupervised} &$88.99$ &$80.34$ &$4.90$ &$1.78$ &$87.71$ &$78.61$ &$\underline{3.78}$ &$1.22$
&$81.30$ &$69.18$ &$\mathbf{17.24}$ &$\underline{4.62}$ &$68.12$ &$54.01$ &$12.71$ &$5.15$
\\
& SimSiam \cite{chen2021exploring} &$88.24$ &$79.18$ &$4.85$ &$1.71$ &$82.71$ &$71.19$ &$5.25$ &$1.78$
&$79.70$ &$67.30$ &$22.83$ &$6.31$ &$51.82$ &$39.11$ &$14.82$ &$6.86$
\\
& SimTriplet \cite{liu2021simtriplet} &$87.99$ &$78.80$ &$5.78$ &$2.00$ &$82.43$ &$70.82$ &$5.24$ &$1.73$
&$80.51$ &$68.24$ &$24.34$ &$7.17$ &$63.55$ &$49.30$ &$13.23$ &$5.98$
\\
& GCL \cite{GCL} &$88.55$ &$79.67$ &$4.22$ &$1.34$ &$\underline{88.01}$ &$\underline{79.04}$ &$4.02$ &$\underline{1.17}$
&$82.41$ &$70.80$ &$22.08$ &$6.59$ &$76.88$ &$63.48$ &$9.11$ &$3.70$
\\
& PCL \cite{zeng2021positional} &$88.97$ &$80.33$ &$4.17$ &$1.43$ &$86.65$ &$76.90$ &$4.23$ &$1.28$
&$\underline{83.13}$ &$\underline{71.70}$ &$20.28$ &$5.17$ &$\underline{78.98}$ &$\underline{66.08}$ &$\underline{8.87}$ &$\underline{3.32}$
\\
& WCL \cite{WCL} &$87.25$ &$77.68$ &$5.33$ &$1.63$ &$84.42$ &$73.66$ &$4.82$ &$1.58$ 
&$81.65$ &$69.75$ &$19.33$ &$5.55$ &$63.87$ &$49.46$ &$11.39$ &$4.70$
\\
& DeSD \cite{DeSD} &$87.14$ &$77.47$ &$6.11$ &$2.19$ &$84.46$ &$73.66$ &$5.15$ &$1.70$
&$81.44$ &$69.44$ &$21.19$ &$5.57$ &$67.96$ &$54.28$ &$11.70$ &$4.90$
\\
& DiRA \cite{DiRA} &$88.92$ &$80.22$ &$4.24$ &$1.48$ &$86.68$ &$77.08$ &$4.46$ &$1.43$ 
&$81.56$ &$69.67$ &$20.85$ &$5.69$ &$64.00$ &$49.87$ &$13.47$ &$5.69$
\\
& \cellcolor{cyan!20!green!20!} SuperCL (Ours) 
&\cellcolor{cyan!20!green!20!}$\mathbf{89.68}$
&\cellcolor{cyan!20!green!20!}$\mathbf{81.44}$ 
&\cellcolor{cyan!20!green!20!}$\mathbf{3.34}$ 
&\cellcolor{cyan!20!green!20!}$\mathbf{1.17}$ 
&\cellcolor{cyan!20!green!20!}$\mathbf{90.26}$ 
&\cellcolor{cyan!20!green!20!}$\mathbf{82.52}$ 
&\cellcolor{cyan!20!green!20!}$\mathbf{3.36}$ 
&\cellcolor{cyan!20!green!20!}$\mathbf{0.98}$ 
&\cellcolor{cyan!20!green!20!}$\mathbf{83.48}$ 
&\cellcolor{cyan!20!green!20!}$\mathbf{72.18}$ 
&\cellcolor{cyan!20!green!20!}$\underline{18.85}$ 
&\cellcolor{cyan!20!green!20!}$\mathbf{4.50}$ 
&\cellcolor{cyan!20!green!20!}$\mathbf{80.02}$ 
&\cellcolor{cyan!20!green!20!}$\mathbf{67.30}$ 
&\cellcolor{cyan!20!green!20!}$\mathbf{7.49}$ 
&\cellcolor{cyan!20!green!20!}$\mathbf{3.03}$
\\
\Xhline{1.0pt} 
\multirow{1}*{100 \%}
& Scratch &$92.07$ &$85.38$ &$3.06$ &$1.07$ &$92.27$ &$85.79$ &$2.71$ &$0.76$
&$84.95$ &$74.33$ &$14.83$ &$3.56$ &$88.65$ &$79.84$ &$4.38$ &$1.62$
\\
\Xhline{1.0pt}
\end{tabular}
\begin{tablenotes}
    \item[1]{The up arrow next to the metric means a larger value is better, while the down arrow means a smaller value is better. The value in \textbf{bold} represents the best and the \underline{underlined} values represent the second best. Scratch means no pre-training.}
  \end{tablenotes}
\end{threeparttable}
\end{table*}

\subsection{Experimental Setup}

In order to verify the superiority of our proposed SuperCL, we conduct comprehensive experiments comparing our SuperCL with other 12 methods on two different types of segmentation tasks: Multi-organ segmentation and ROI-based (region-of-interest, such as tumor, skin lesion, etc.) segmentation. We use 3 public avaliable medical image datasets: CHD (17525 slices), BraTS2018 (39064 slices) and KiTS2019 (32332 slices) for pre-training and 8 datasets for fine-tuning and evaluating downstream segmentation performance.

\noindent{\textbf{Upstream Datasets:}}
\begin{enumerate}
    \item {\itshape {The CHD dataset}} \cite{PCL23} is a CT dataset that consists of 68 3D cardiac images captured by a Simens biograph 64 machine. The segmentation labels include seven substructures: left ventricle (LV), right ventricle (RV), left atrium (LA), right atrium (RA), myocardium (Myo), aorta (Ao) and pulmonary artery (PA).
    \item {\itshape The BraTS2018 dataset} \cite{bakas2018identifying} contains a total of 351 patients scanned with T1, T1CE, T2, and Flair MRI volumes. 
    Notably, since there are no suitable MRI datasets aimed for multi-organ segmentation, we use the BraTS2018 dataset as the pre-training dataset for both multi-organ and ROI-based MRI segmentation downstream tasks.
    \item {\itshape The KiTS2019 dataset} \cite{heller2021state} consists of 210 patients, aimed for kidney and kidney tumor segmentation and we choose it as the pre-training dataset for ROI-based CT segmentation.
\end{enumerate}

\noindent{\textbf{Downstream Multi-organ Segmentation Datasets:}}
\begin{enumerate}
    \item {\itshape The ACDC dataset} \cite{ACDC} has 100 patients with 3D cardiac MRI images and the segmentation labels include three substructures: LV, RV and Myo. 
    \item {\itshape The MMWHS dataset} \cite{PCL24} consists of 20 cardiac CT and 20 MRI images and the annotations include the same seven substructures as the CHD dataset. And we use the 20 cardiac CT images for fine-tuning. 
    \item {\itshape The HVSMR dataset} \cite{PCL16} has 10 3D cardiac MRI images captured in an axial view on a 1.5T scanner. Manual annotations of blood pool and Myo are provided.
    \item {\itshape The CHAOS dataset} \cite{CHAOS2021} has totally 120 DICOM datasets from T1-DUAL in/out phase and T2-SPIR. And we choose T2-SPIR train dataset with 20 DICOM MRI images and corresponding labels of four regions: liver, left kidney, right kidney and spleen for fine-tuning.
\end{enumerate}

\noindent{\textbf{Downstream ROI-based Segmentation Datasets:}}
\begin{enumerate}
    \item {\itshape The MSD dataset} \cite{antonelli2022medical} has 10 avaliable datasets, where each dataset has between one and three region-of-interest (ROI) targets. And we use three datasets: 
    {Heart} (Task02, MRI dataset, 20 patients labeled with left atrium), {Hippocampus} (Task04, MRI dataset, 260 patients labeled with anterior and posterior of hippocampus) and {Spleen} (Task09, CT datset, 41 patients labeled with spleen).
    \item {\itshape The ISIC2018 dataset} \cite{DBLP:journals/corr/abs-1902-03368} is a comprehensive collection of dermoscopic images published by the International Skin Imaging Collaboration (ISIC). And for the lesion segmentation task, the dataset includes 2594 images with corresponding annotations of lesion skin region which are reviewed by dermatologists.
\end{enumerate}


\begin{table*}[th]
\centering
\caption{Quantitative results of our proposed SuperCL against other SOTA methods on 4 ROI-based segmentation datasets.}
\scriptsize
\setlength{\tabcolsep}{1pt}
\begin{tabular}{c|c|cccc|cccc|cccc|cccc}
\Xhline{1.0pt} 
 &  & \multicolumn{4}{c|}{ Spleen (41 patients) } & \multicolumn{4}{c|}{ ISIC (2594 images) } & \multicolumn{4}{c|}{ Heart (20 patients) } & \multicolumn{4}{c}{ Hippocampus (260 patients) }\\
 {Labeled}& {Method} & DSC (\%)$\uparrow$& JC (\%)$\uparrow$& HD95$\downarrow$& ASD$\downarrow$& DSC (\%)$\uparrow$& JC (\%)$\uparrow$& HD95$\downarrow$& ASD$\downarrow$
 & DSC (\%) $\uparrow$ & JC (\%) $\uparrow$ & HD95 $\downarrow$ & ASD $\downarrow$
 & DSC (\%)$\uparrow$& JC (\%)$\uparrow$& HD95$\downarrow$& ASD $\downarrow$
 \\

\Xhline{1.0pt} \multirow{12}*{10 \%}
& Scratch &$66.63$ &$49.95$ &$12.58$ &$5.78$&$\underline{87.21}$&$\underline{77.32}$&$23.47$&$9.72$ 
&$73.30$ &$57.85$ &$5.21$ &$1.82$ &$\underline{78.27}$ &$\underline{64.30}$ &$1.36$ &$0.51$
\\
& SimCLR \cite{chen2020simple} &$60.38$ &$43.24$ &$12.42$ &$4.67$&$85.85$&$75.21$&$23.81$&$10.32$ 
&$73.42$&$58.00$&$\mathbf{4.32}$&$\underline{1.40}$ &$75.91$ &$61.18$ &$1.48$ &$0.54$
\\
& MoCo \cite{he2020momentum} &$56.49$ &$39.36$ &$21.32$ &$9.35$&$85.61$&$74.84$&$24.56$&$9.95$
&$68.32$&$51.88$&$6.71$&$2.48$ &$78.27$ &$64.30$ &${\underline{1.27}}$ &$0.46$
\\
& BYOL \cite{grill2020bootstrap} &$61.38$ &$44.28$ &$14.56$ &$6.32$&$87.00$&$77.00$&$24.20$&$10.28$ 
&$70.34$ &$58.00$ &${{6.17}}$ &$2.16$ &$77.24$ &$62.92$ &$1.39$ &$0.49$
\\
& SwAV \cite{caron2020unsupervised} &$61.87$ &$44.79$ &$13.61$ &$6.53$&$85.85$&$75.20$&$24.19$&$10.45$ 
&$72.31$ &$56.63$ &$5.95$ &$2.27$ &$76.00$ &$61.29$ &$1.47$ &$0.52$
\\
& SimSiam \cite{chen2021exploring} &$67.88$ &$51.37$ &$12.89$ &$3.65$&$83.02$&$70.97$&$30.04$&$13.70$
&$58.56$ &$41.41$ &$10.71$ &$5.15$ &$73.45$ &$58.05$ &$1.39$ &$0.49$
\\
& SimTriplet \cite{liu2021simtriplet} &$63.56$ &$46.59$ &$16.95$ &$7.54$&$83.01$&$70.96$&$29.27$&$13.03$ 
&$57.76$ &$40.61$ &$8.85$ &$3.94$ &$74.80$ &$59.74$ &$1.31$ &$\underline{0.46}$
\\
& GCL \cite{GCL} &$\underline{73.40}$ &$\underline{57.97}$ &$11.79$ &$4.99$&$85.53$&$74.72$&$24.37$&$10.84$
&$76.88$&$62.44$&${5.21}$&$1.73$&$78.14$ &$64.13$ &$1.32$ &$0.46$
\\
& PCL \cite{zeng2021positional} &${71.22}$ &$55.30$ &$\underline{11.10}$ &${4.97}$&$86.64$&$76.43$&$24.41$&$10.50$
&$\underline{78.47}$ &$\underline{64.57}$ &$5.95$ &$2.14$ &$78.13$ &$64.11$ &${1.29}$ &$0.47$
\\
& WCL \cite{WCL} &$64.75$ & $47.88$ & $16.70$& $5.46$&$86.30$&$75.90$&$23.13$&$9.77$ 
&$74.14$ &$58.91$ &$6.13$ &$1.93$ &$76.57$ &$62.03$ &$1.36$ &$0.46$
\\
& DeSD \cite{DeSD} &$64.24$ &$47.31$ &$\mathbf{7.18}$ &${\mathbf{2.83}}$&$83.56$&$71.76$&$25.66$&$11.50$
&$77.55$ &$63.33$ &$5.80$ &$2.28$ &$74.48$ &$59.34$ &$1.64$ &$0.62$
\\
& DiRA \cite{DiRA} &$66.23$ &$49.51$ &$20.96$ &$8.93$&$87.02$&$77.02$&$\underline{22.91}$&$\underline{9.28}$
&$70.19$&$54.07$ &$\underline{4.76}$&${\mathbf{1.36}}$&$77.15$ &$62.80$ &$1.36$ &$0.50$
\\
&\cellcolor{cyan!20!green!20!}SuperCL (Ours) 
&\cellcolor{cyan!20!green!20!}$\mathbf{81.29}$ 
&\cellcolor{cyan!20!green!20!}$\mathbf{68.48}$ 
&\cellcolor{cyan!20!green!20!}$11.28$ 
&\cellcolor{cyan!20!green!20!}$\underline{4.34}$ 
&\cellcolor{cyan!20!green!20!}$\mathbf{88.44}$ 
&\cellcolor{cyan!20!green!20!}$\mathbf{79.28}$ 
&\cellcolor{cyan!20!green!20!}$\mathbf{21.33}$ 
&\cellcolor{cyan!20!green!20!}$\mathbf{8.90}$ 
&\cellcolor{cyan!20!green!20!}$\mathbf{78.63}$ 
&\cellcolor{cyan!20!green!20!}$\mathbf{64.79}$ 
&\cellcolor{cyan!20!green!20!}$4.96$ 
&\cellcolor{cyan!20!green!20!}$1.73$ 
&\cellcolor{cyan!20!green!20!}$\mathbf{80.51}$ 
&\cellcolor{cyan!20!green!20!}$\mathbf{67.37}$ 
&\cellcolor{cyan!20!green!20!}$\mathbf{1.18}$ 
&\cellcolor{cyan!20!green!20!}$\mathbf{0.42}$
\\
\Xhline{1.0pt} \multirow{12}*{25 \%}
& Scratch&$74.99$ &$59.98$ &$9.14$ &$4.33$&$88.44$&$79.28$&${\mathbf{16.32}}$&${\mathbf{6.41}}$
&$78.35$&$64.41$ &$4.57$ &$1.33$ &$83.48$ &$71.65$ &$1.08$ &$0.35$
\\
& SimCLR \cite{chen2020simple} &$81.75$ &$69.13$ &$7.05$ &$3.06$&$87.82$&$78.28$&$17.79$&$7.18$
&$80.07$ &$66.76$ &${\mathbf{3.48}}$ &$0.98$ &$83.35$ &$71.46$ &$1.05$ &$0.36$
\\
& MoCo \cite{he2020momentum} &$75.98$ &$61.26$ &$8.55$ &$4.09$&$87.53$&$77.83$&$17.87$&$7.01$
&$75.35$&$60.45$&$5.60$&$2.11$ &$83.94$ &$72.33$ &${0.98}$ &$\underline{0.33}$
\\
& BYOL \cite{grill2020bootstrap} &$75.15$ &$60.19$ &$6.34$ &$2.39$&$88.50$&$79.38$&$\underline{16.78}$&$\underline{6.54}$
&$80.70$ &$67.65$ &$4.36$ &$1.16$ &$83.38$ &$71.50$ &$1.04$ &$0.35$
\\
& SwAV \cite{caron2020unsupervised} &$80.40$ &$67.23$ &$\underline{5.52}$ &$\underline{2.34}$&$87.90$&$78.41$&$17.65$&$6.77$
&$81.94$ &$69.40$ &$3.98$ &$1.22$ &$83.05$ &$71.02$ &$1.06$ &$0.35$
\\
& SimSiam \cite{chen2021exploring} &$78.70$ &$64.88$ &$8.87$ &$4.56$&$87.29$&$77.45$&$19.31$&$7.52$
&$71.65$ &$55.83$ &$3.74$ &$1.17$ &$80.60$ &$67.51$ &$1.11$ &$0.37$
\\
& SimTriplet \cite{liu2021simtriplet} &$78.81$ &$65.04$ &$8.84$ &$3.96$&$86.52$&$76.24$&$19.50$&$7.68$
&$76.42$ &$61.83$ &$5.11$ &$1.72$ &$82.01$ &$69.51$ &$1.06$ &$0.35$
\\
& GCL \cite{GCL} &$\underline{84.37}$ &$\underline{72.97}$ &${6.25}$ &${2.52}$&$\underline{88.52}$&$\underline{79.41}$&$17.55$&$7.06$
&$82.30$ &$69.92$ &$3.76$ &${0.96}$ &$83.63$ &$71.87$ &$1.07$ &$0.37$
\\
& PCL \cite{zeng2021positional} &$83.55$ &$71.74$ &$7.03$ &$2.96$&$88.26$&$78.99$&$18.04$&$7.27$
&$\underline{82.69}$&$\underline{70.49}$&$3.78$&${\underline{0.93}}$ &$\underline{83.98}$ &$\underline{72.40}$ &${\underline{0.96}}$ &$0.33$
\\
& WCL \cite{WCL} & $78.27$ & $64.30$ & $6.94$ & $2.56$&$87.88$&$78.38$&$16.47$&$6.59$ 
&$78.44$ &$64.52$ &$4.38$ &$1.21$ &$83.78$ &$72.09$ &$1.04$ &$0.36$
\\
& DeSD \cite{DeSD} &$81.02$ &$68.10$ &$6.59$ &$3.28$&$87.34$&$77.52$&$20.29$&$8.41$
&$81.50$ &$68.77$ &$4.45$ &$1.43$ &$81.81$ &$69.22$ &$1.08$ &$0.37$
\\
& DiRA \cite{DiRA} &$80.80$ &$67.78$ &$7.95$ &$4.17$&$88.38$&$79.18$&$16.96$&$6.63$ 
&$79.91$ &$66.54$ &$4.87$ &$1.35$ &$83.60$ &$71.83$ &$1.00$ &$0.33$
\\
&\cellcolor{cyan!20!green!20!}SuperCL (Ours) 
&\cellcolor{cyan!20!green!20!}$\mathbf{88.49}$ 
&\cellcolor{cyan!20!green!20!}$\mathbf{79.36}$ 
&\cellcolor{cyan!20!green!20!}$\mathbf{4.89}$ 
&\cellcolor{cyan!20!green!20!}$\mathbf{2.10}$ 
&\cellcolor{cyan!20!green!20!}$\mathbf{89.56}$ 
&\cellcolor{cyan!20!green!20!}$\mathbf{81.09}$ 
&\cellcolor{cyan!20!green!20!}$17.18$ 
&\cellcolor{cyan!20!green!20!}$6.70$ 
&\cellcolor{cyan!20!green!20!}$\mathbf{83.98}$ 
&\cellcolor{cyan!20!green!20!}$\mathbf{72.38}$ 
&\cellcolor{cyan!20!green!20!}$\underline{3.56}$ 
&\cellcolor{cyan!20!green!20!}$\mathbf{0.89}$ 
&\cellcolor{cyan!20!green!20!}$\mathbf{85.32}$ 
&\cellcolor{cyan!20!green!20!}$\mathbf{74.40}$ 
&\cellcolor{cyan!20!green!20!}$\mathbf{0.95}$ 
&\cellcolor{cyan!20!green!20!}$\mathbf{0.32}$
\\
\Xhline{1.0pt} \multirow{1}*{100 \%}
& Scratch &$89.92$ &$81.69$ &$4.59$ &$2.12$&$90.11$&$82.01$&$15.71$&$6.17$ 
&$91.52$ &$84.36$ &$2.45$ &$0.61$ &$87.00$ &$77.00$ &$0.81$ &$0.27$
\\
\Xhline{1.0pt}
\end{tabular}
\label{ROI-based}
\end{table*}

\noindent{\textbf{Implementation Details:}} 
We employ our SuperCL to pre-train a U-Net encoder on CHD, BraTS and KiTS dataset, respectively without using any human label. Then the pre-trained model is used as the initialization to fine-tune a UNet segmentation network (we add the decoder to form the whole UNet). 
All the downstream 8 datasets are split into train set and test set in an 8:2 ratio. And except for Scratch with 100\% annotations (fully supervised learning without pre-training) is used as the upper bound, we only use a small fraction (10\% / 25\%) annotations from train set of 8 downstream datasets to finetune the UNet in our SuperCL and other CL methods. Four metrics including DSC, JC, HD95 and ASD are used to evaluate the segmentation performance.
Experiments of pre-training and fine-tuning are conducted with PyTorch on NVIDIA A100 80G GPUs. Non-spatial augmentations are brightness, contrast, gaussian blur, and spatial augmentations are translation, rotation and scale in the pre-training stage and the data augmentations in the fine-tuning stage are translation, rotation and scale. In the pre-training stage, the weighted terms $\{\lambda_1, \lambda_2, \lambda_3\}$ are empirically set to be $\{1.0, 1.0, 0.5\}$, temperature $\tau$ is set to be 0.1, the model is trained with 100 epochs, 16 batches per GPU, SGD optimizer and initial learning rate of 0.1, which is then decayed to 0 with the cosine scheduler on each training iterator. In the fine-tuning stage, we train the U-Net with cross-entropy loss for 100 epochs with Adam optimizer. The batch size per GPU is set to be 5, and initial learning rate is $5\times e^{-4}$ which is decayed with cosine scheduler to minimum learning rate $5 \times e^{-6}$.

\section{Results and Discussions}


\subsection{Results}


\noindent{\textbf{Comparison Studies:}} 
We compare the performance of our SuperCL with the scratch approach which does not use any pre-training strategy as well as other state-of-the-art CL baselines (CL baselines are pre-trained with each upstream dataset, the same as our SuperCL), including (1) SimCLR \cite{chen2020simple}, MoCo \cite{he2020momentum}, BYOL \cite{grill2020bootstrap}, SwAV \cite{caron2020unsupervised}, SimSiam \cite{chen2021exploring} and  WCL \cite{WCL} originally used in natural imaging domain but modified with an encoder of UNet as the backbone of CL; (2) SimTriplet \cite{liu2021simtriplet}, GCL \cite{GCL}, PCL \cite{zeng2021positional}, DeSD \cite{DeSD}, DiRA \cite{DiRA} specially used in medical imaging domain. All the experiments across different methods have the same dataset settings and partition.
and all the methods have the same backbone structures. 
To make it clear, we summarize our comparative studies as ``upstream dataset (pre-training) $\rightarrow$ downstream dataset (fine-tuning)": for multi-organ segmentation task, it is CHD (CT) $\rightarrow$ ACDC/MMWHS, BraTS (MRI) $\rightarrow$ HVSMR/CHAOS; while for ROI-based segmentation task, it is KiTS (CT) $\rightarrow$ Spleen/ISIC, BraTS (MRI) $\rightarrow$ Heart/Hippocampus. 
The quantitative results carried on multi-organ segmentation task are shown in Table \ref{mutiorgan} while ROI-based segmentation task in Table \ref{ROI-based}.

\begin{table*}[t]
\centering
\caption{The ablation studies for each component of our proposed SuperCL.}
$\begin{aligned}
\begin{tabular}{lcc|cccc|cccc}
\Xhline{1.0pt}
\multicolumn{3}{c|}{\text { Settings }} & \multicolumn{4}{c|}{\text { ACDC (25\% label) }} & \multicolumn{4}{c}{\text {MMWHS (25\% label)}} \\
Framework & ILCP & IGCP & DSC (\%)$\uparrow$& JC (\%)$\uparrow$& HD95$\downarrow$& ASD$\downarrow$& DSC (\%)$\uparrow$& JC (\%)$\uparrow$& HD95$\downarrow$& ASD$\downarrow$\\
\Xhline{1.0pt}
WCL (Baseline 1) &$\times$& $\times$ &$87.25$ &$77.68$ &$5.33$ &$1.63$ &$84.42$ &$73.66$ &$4.82$ &$1.58$ \\
WCL + IGCP & $\times$ & $\checkmark$ &$88.23$&$79.16$& $5.05$& $1.83$ &$84.60$ &$73.98$ &$4.53$ &$1.45$ \\
WCL + ILCP & $\checkmark$ & $\times$&$88.90$&$80.20$&$4.34$&$1.41$&$87.70$ &$78.57$ &$4.44$ &$1.34$ \\
\cellcolor{cyan!20!green!20!}WCL + IGCP + ILCP & \cellcolor{cyan!20!green!20!}$\checkmark$ & \cellcolor{cyan!20!green!20!}$\checkmark$ &\cellcolor{cyan!20!green!20!}$\mathbf{89.30}$ &\cellcolor{cyan!20!green!20!}$\mathbf{80.83}$ &\cellcolor{cyan!20!green!20!}$\mathbf{3.70}$ &\cellcolor{cyan!20!green!20!}$\mathbf{1.22}$ &\cellcolor{cyan!20!green!20!}$\mathbf{89.13}$ &\cellcolor{cyan!20!green!20!}$\mathbf{80.71}$ &\cellcolor{cyan!20!green!20!}$\mathbf{3.95}$ &\cellcolor{cyan!20!green!20!}$\mathbf{1.22}$  \\
\Xhline{1.0pt}
PCL (Baseline 2) & $\times$& $\times$&$88.97$&$80.33$&$4.17$&$1.43$&$86.65$ &$76.90$ &$4.23$ &$1.28$ \\
PCL + IGCP &$\times$ & $\checkmark$ &$89.01$&$80.38$&$5.31$&$2.03$&$88.51$&$79.80$&$3.85$&$1.15$ \\
PCL + ILCP &$\checkmark$ &$\times$
&$89.11$&$80.50$&$3.75$&$1.20$&$88.64$&$79.94$&$3.97$&$1.19$ \\
$\cellcolor{cyan!20!green!20!}$PCL + IGCP + ILCP &$\cellcolor{cyan!20!green!20!}$$\checkmark$&$\cellcolor{cyan!20!green!20!}$ $\checkmark$
&$\cellcolor{cyan!20!green!20!}$$\mathbf{89.68}$
&$\cellcolor{cyan!20!green!20!}$$\mathbf{81.07}$ 
&$\cellcolor{cyan!20!green!20!}$$\mathbf{3.36}$ 
&$\cellcolor{cyan!20!green!20!}$$\mathbf{1.17}$ 
&$\cellcolor{cyan!20!green!20!}$$\mathbf{90.26}$ 
&$\cellcolor{cyan!20!green!20!}$$\mathbf{82.52}$ 
&$\cellcolor{cyan!20!green!20!}$$\mathbf{3.36}$ 
&$\cellcolor{cyan!20!green!20!}$$\mathbf{0.98}$ \\
\Xhline{1.0pt}
\end{tabular}
\end{aligned}$
\label{Ablation}
\end{table*}

\begin{figure*}[t]
  \centering
  \includegraphics[width=0.9\textwidth]{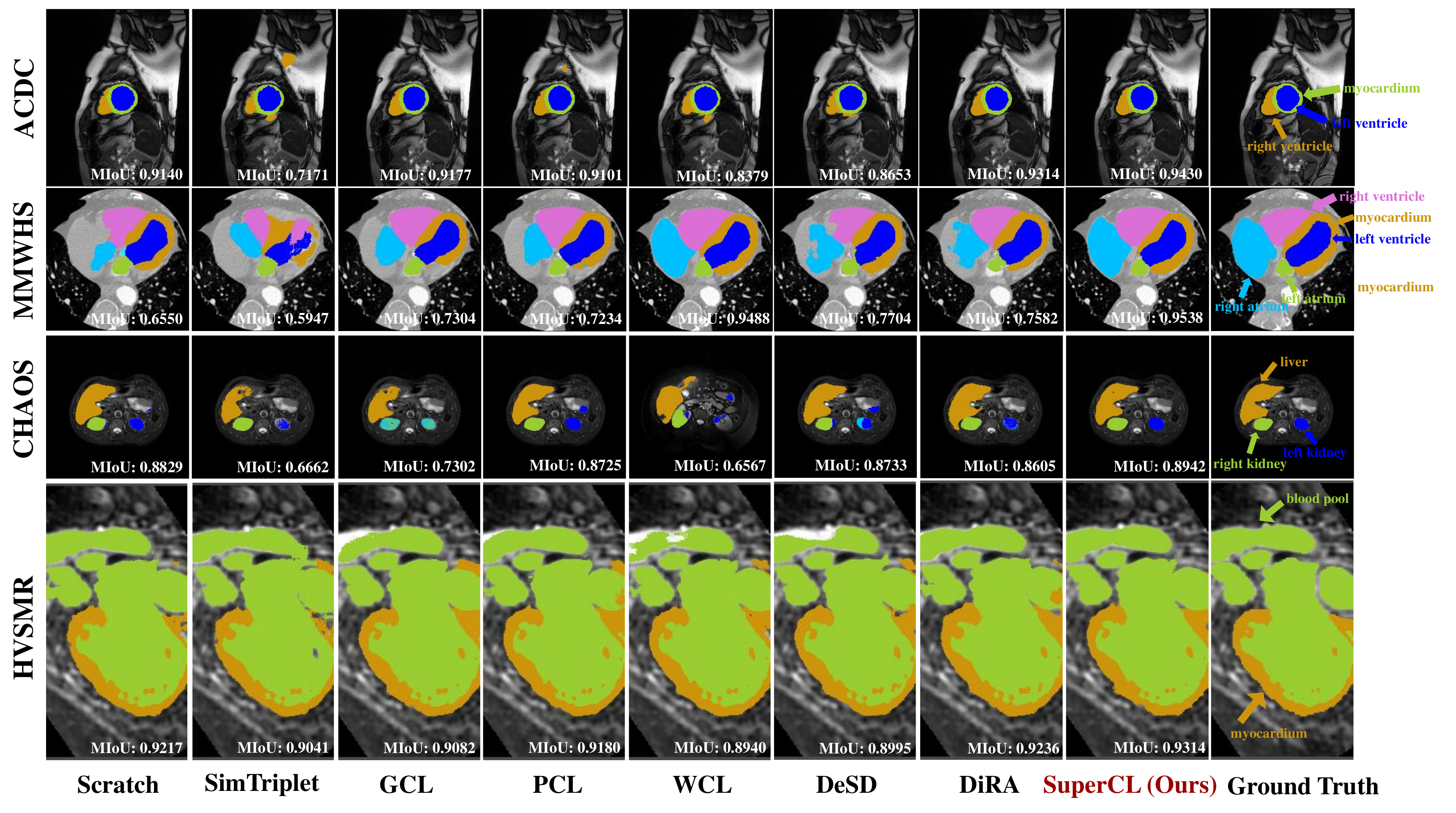}
  \caption{Visualization of multi-organ segmentation results on ACDC, MMWHS, CHAOS and HVSMR.}
  \label{vis_1}
\end{figure*}

\noindent{{\textbf{Quantitative results:}}} 
From Table \ref{mutiorgan} and \ref{ROI-based}, we have the following observations: In general, benefited from ILCP and IGCP for contrastive pairs generation, our SuperCL outperforms all the existing SOTA baselines in almost all the metrics under different ratios of labeled data. Specifically, (1) when 10\% annotations are used, our proposed SuperCL gains a great improvement with 
3.98\% DSC and 4.90\% JC higher than the previous best baseline on average across 8 datasets.
(2) When 25\% annotations are used, the performance improvement of our SuperCL against other SOTA baselines is narrowed. Because with more labeled samples, more supervisory information will be available, which results in that the improvement from self-supervised pre-training tends to saturate. Despite that, the gain of performance is still stable with 2.89\% DSC and 3.54\% JC higher than the previous best baseline on average across 8 datasets. 
(3) Moreover, our proposed SuperCL pre-training framework can realize our initial goal effectively: pre-train a model with unlabeled data from upstream datasets and limited labeled data from downstream datasets to achieve comparative performance with fully-supervised trained model. Specifically, the performance of our SuperCL with 25\% annotations can be paralleled with that of Scratch with 100\% annotations (fully-supervised training). 
{\itshape e.g.} 90.26\% vs. 92.27\% DSC on MMWHS; 88.49\% vs. 89.92\% DSC on Spleen; 89.56\% vs. 90.11\% DSC on ISIC.  

\noindent{{\textbf{Visualization of segmentation results:}}} 
As shown in Figure \ref{vis_1}, we visualize the exemplar qualitative results of our SuperCL and other 7 methods (more visualization results are shown in Appendix Fig.\ref{multi_organ_1}) on ACDC, MMWHS, CHAOS and HVSMR.
Compared with different methods, it can be seen that our proposed SuperCL achieves a superior performance with a higher MIoU and more precise predictions of substructures across other 7 methods. {\itshape e.g.} Our SuperCL achieves MIoU 0.9430 while SimTriplet only gets MIoU 0.7171 with some wrong prediction region (right ventricle/yellow region), same as PCL, WCL and DeSD. And almost all the SOTA baselines including GCL, PCL, DeSD, DiRA have a large missing segmentation region (right atrium/sky-blue region) with a much lower MIoU while our proposed SuperCL has an impressive segmentation result with MIoU 0.9538 on MMWHS. 4 ROI-based segmentation datasets in Appendix Fig. \ref{ROI_1}. Notably, because there is usually one class label in ROI-based segmentation dataset and the label region is relatively the minority of the whole image, therefore, we appropriately enlarge the visualization results of Spleen, Heart and ISIC to better observe the difference of segmentation performance between our SuperCL and other baseline methods. The locations of ROIs in the original images can be found in the first row of Fig. \ref{ROI_1}. Compared with different baseline methods, it can be seen that our proposed SuperCL achieves a superior performance with a higher MIoU and more precise predictions of substructures across other 12 methods.

\noindent{{\textbf{Ablation Studies:}}} 
We perform an ablation study with WCL and PCL as two baselines to investigate the effects of the two key components: ILCP and IGCP of our proposed SuperCL. As shown in Table \ref{Ablation}, we conclude that both strategies contribute to the improvement of segmentation performance and ILCP matters more and improves the performance to a greater extent than IGCP. Specifically, (1) $\mathbf{X}$ + IGCP outperforms two baselines, which means our proposed ASPCCL is effective. 
{\itshape e.g.} 1.86\% DSC higher than PCL on MMWHS. (2) $\mathbf{X}$ + ILCP also outperforms two baselines and even $\mathbf{X}$ + IGCP. This is mainly because ILCP exploits the pixel-level prior knowledge and correlation which contributes a lot the downstream segmentation task which focuses on intra-image discrimination. {\itshape e.g.} 0.67\% DSC and 2.52\% JC promotion than WCL on ACDC and 3.10\% DSC and 4.59\% JC promotion than WCL + IGCP on MMWHS. (3) Finally, when combining the ILCP and IGCP together ($\mathbf{X}$ + IGCP + ILCP), our complete SuperCL achieves the best segmentation performace, which means with the cooperation of ILCP and IGCP, our proposed SuperCL can make a much greater improvement than either single one strategy. And PCL + IGCP +ICLP is better, which is also used as our experimental setting. 

\subsection{Discussions}

\begin{figure}[t]
  \centering
  \includegraphics[width=0.5\textwidth]{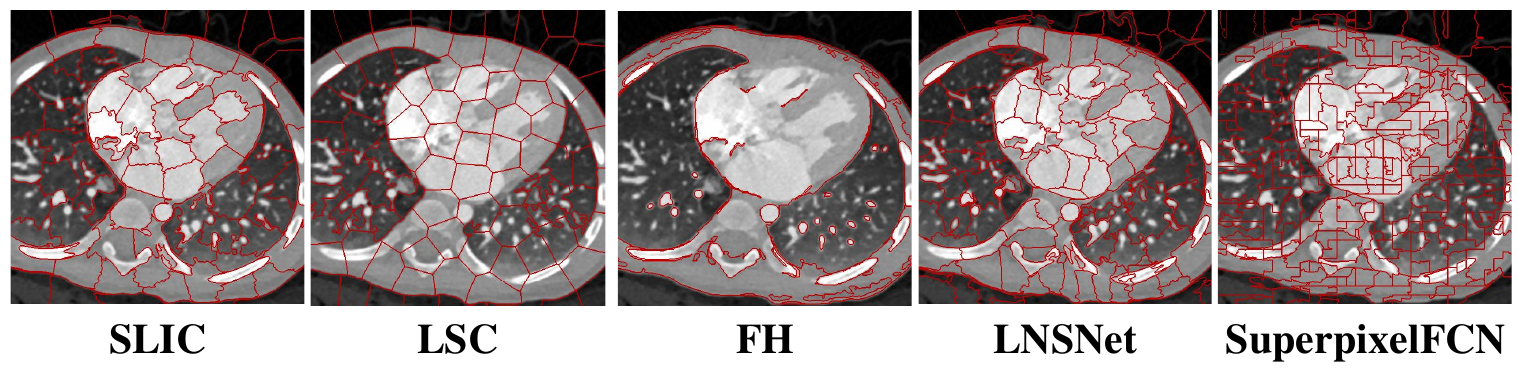}
  \includegraphics[width=0.5\textwidth]{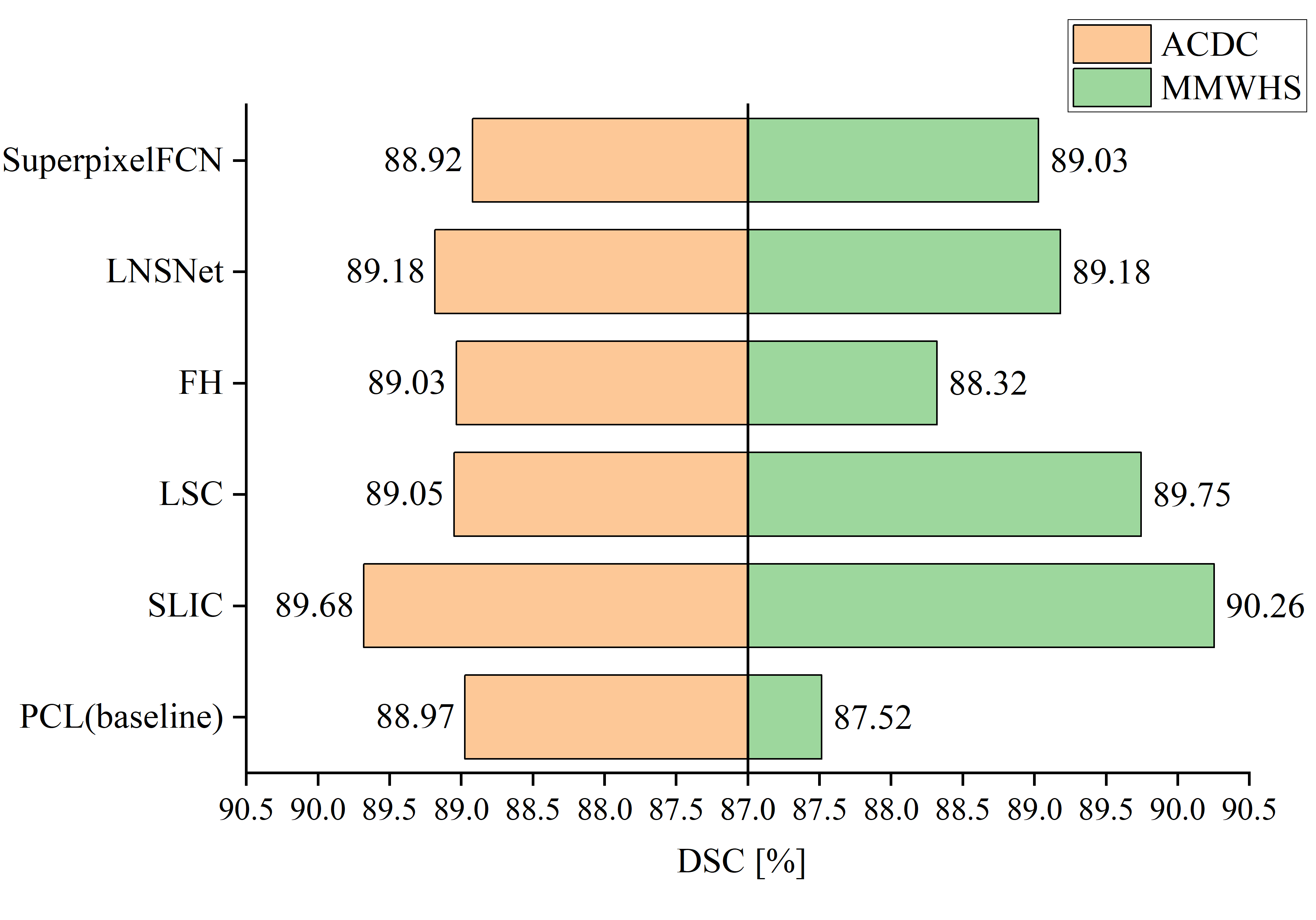}
  \caption{Comparison results of different superpixel methods on ACDC and MMWHS with 25\% annotations.}
  \label{sp_method}
\end{figure}

\noindent{\textbf{Different superpixel methods:}} We make a comparison experiment between different superpixel methods, including cluster-based methods: SLIC \cite{SLIC} and LSC \cite{LSC}; graph-based method: FH \cite{FH} and CNN-based methods: LNSNet \cite{LNSNet} and SuperpixelFCN \cite{SuperpixelFCN}. According to Fig. \ref{sp_method}, we conclude that the performance improvement depends heavily on the shape and structure of the superpixel pseudo mask. And cluster-based methods SLIC and LSC are generally better than graph-based method FH and CNN-based method SuperpixelFCN. 
Resonably, the concept of clustering aligns well with the idea of dividing positive and negative sample pairs in CL, making it more suitable for CL strategy. As for CNN-based methods, according to the visualization of the superpixel pseudo mask, LNSNet generates a much more reliable and similar mask as SLIC, compared with SuperpixelFCN, which hence achieves a paralleled performance (even outperforms LSC on ACDC). In terms of graph-based method FH, it focuses more on the edge details, which may be helpful to segmentation directly but less effective for constructing positive and negative pairs for CL. However, almost all the superpixel methods perform superior to PCL baseline (except for superpixelFCN on ACDC), which testifies the effectiveness and versatility of our proposed SuperCL.

\begin{table}[t]
  \centering
  \scriptsize
  \setlength{\tabcolsep}{5pt}
  \caption{The sensitivity studies of stride of SuperCL (w/o $\mathcal{L}_{inter}$).}
  \begin{tabular}{cc|c|cc|cc}
  \hline 
  & & & \multicolumn{2}{c|}{ACDC (25\% label)} & \multicolumn{2}{c}{MMWHS (25\% label)} \\
  Methods &  Stride &  Pixels & DSC (\%)$\uparrow$& HD95$\downarrow$& DSC (\%)$\uparrow$& HD95$\downarrow$\\
  \hline
  PCL (baseline) & -& -& $88.97$ &$4.17$&$87.52$&$4.25$\\
  \hline
  \multirow[]{7}{*}{SuperCL} &$\cellcolor{cyan!20!green!20!}$1 &$\cellcolor{cyan!20!green!20!}$1024& $\cellcolor{cyan!20!green!20!}$$\mathbf{89.68}$ &$\cellcolor{cyan!20!green!20!}$$\mathbf{3.34}$& $\cellcolor{cyan!20!green!20!}$$\mathbf{90.26}$ &$\cellcolor{cyan!20!green!20!}$$\mathbf{3.36}$ \\
 \multirow[]{7}{*}{(w/o $\mathcal{L}_{inter}$)}&2 &512&$89.04$ &$4.03$&  $88.48$ &$4.00$  \\
&4 &256&$88.91$ &$3.66$  &$88.90$ &$3.49$ \\
&8&128&$89.10$ &$3.63$  &$88.89$ &$3.65$ \\
&16 &64&$88.98$ &$3.82$  &$90.09$ &$3.40$  \\
&32&32&$88.94$ &$3.79$  &$89.22$ &$3.71$ \\
&64&16&$88.97$ &$3.55$  &$89.00$ &$3.76$  \\
  \hline
  \end{tabular}
\label{stride}
\end{table}

\begin{table}[t]
\centering
\caption{Quantitative results on ACDC with 10\% and 20\% annotations of our SuperCL in semi-supervised learning scenario.}
\scriptsize
\setlength{\tabcolsep}{3pt}
\begin{tabular}{c|c|cccc}
\Xhline{1.0pt} 
{Labeled\%}& {Method} & DSC (\%) $\uparrow$ & JC (\%) $\uparrow$ & HD95 $\downarrow$ & ASD $\downarrow$ \\
\Xhline{1.0pt}
\multirow{14}*{10 \%}
&SL & $77.66$ & $66.40$ & $2.41$ & $0.59$ \\
&MT & $81.11$ & $69.99$ & $8.99$ & $2.70$ \\
&UA-MT & $80.71$ & $69.58$ & $13.69$ & $4.50$ \\
&SASSNet & $82.56$ & $71.90$ & $9.13$ & $2.64$ \\
&DTC & $84.32$ & $74.04$ & $9.47$ & $2.63$ \\
&URPC & $82.41$ & $71.69$ & $5.83$ & $1.65$ \\
&CPS & $84.24$ & $73.91$ & $8.26$ & $2.45$ \\
&CPSCauSSL & $85.25$ & $75.31$ & $6.05$ & $1.97$ \\
&MC-Net+ & $86.14$ & $76.61$ & $6.04$ & $1.85$ \\
&MCCauSSL & $86.80$ & $77.48$ & $5.73$ & $1.83$ \\
&BCP & $88.48$ & $80.62$ & $3.98$ & $1.17$ \\
&$\cellcolor{cyan!20!green!20!}$SuperCL + BCP &$\cellcolor{cyan!20!green!20!}$$\mathbf{89.71}$ &$\cellcolor{cyan!20!green!20!}$$\mathbf{81.91}$ &$\cellcolor{cyan!20!green!20!}$$\mathbf{1.77}$ &$\cellcolor{cyan!20!green!20!}$$\mathbf{0.70}$\\
&BCPCauSSL & $89.66$ & $81.79$ & $3.67$ & $0.93$ \\
& $\cellcolor{cyan!20!green!20!}$SuperCL + BCPCauSSL &$\cellcolor{cyan!20!green!20!}$$\mathbf{89.74}$ &$\cellcolor{cyan!20!green!20!}$$\mathbf{81.90}$ &$\cellcolor{cyan!20!green!20!}$$\mathbf{2.09}$ & $\cellcolor{cyan!20!green!20!}$$\mathbf{0.75}$ \\
\Xhline{1.0pt}
\multirow{14}*{20 \%}
&SL & $84.62$ & $74.85$ & $6.32$ & $1.79$ \\
&MT & $85.46$ & $75.89$ & $8.02$ & $2.39$ \\
&UA-MT & $85.16$ & $75.49$ & $5.91$ & $1.79$ \\
&SASSNet & $86.45$ & $77.20$ & $6.63$ & $1.98$ \\
&DTC & $87.10$ & $78.15$ & $6.76$ & $1.99$ \\
&URPC & $85.44$ & $76.36$ & $5.93$ & $1.70$ \\
&CPS & $86.85$ & $77.96$ & $5.48$ & $1.64$ \\
&CPSCauSSL & $87.24$ & $78.44$ & $5.57$ & $1.73$ \\
&MC-Net+ & $87.10$ & $78.21$ & $5.04$ & $1.56$ \\
&MCCauSSL & $87.84$ & $79.32$ & $4.37$ & $1.28$ \\
&BCP & $89.52$ & $81.62$ & $3.69$ & $1.03$ \\
&$\cellcolor{cyan!20!green!20!}$SuperCL + BCP & $\cellcolor{cyan!20!green!20!}$$\mathbf{89.93}$ & $\cellcolor{cyan!20!green!20!}$$\mathbf{82.31}$ & $\cellcolor{cyan!20!green!20!}$$\mathbf{3.34}$ & $\cellcolor{cyan!20!green!20!}$$\mathbf{0.92}$ \\
&BCPCauSSL & $89.99$ & $82.34$ & $3.60$ & $0.88$ \\
&$\cellcolor{cyan!20!green!20!}$SuperCL + BCPCauSSL & $\cellcolor{cyan!20!green!20!}$$\mathbf{90.35}$ & $\cellcolor{cyan!20!green!20!}$$\mathbf{82.89}$ & $\cellcolor{cyan!20!green!20!}$$\mathbf{1.64}$ & $\cellcolor{cyan!20!green!20!}$$\mathbf{0.50}$ \\
\Xhline{1.0pt}
\multirow{1}*{100 \%}
&SL (upper bound) & $91.53$ & $84.76$ & $2.41$ & $0.59$ \\
\Xhline{1.0pt}
\end{tabular}
\label{SemiSL}
\end{table}

\begin{figure}[t]
  \centering
  \includegraphics[width=0.5\textwidth]{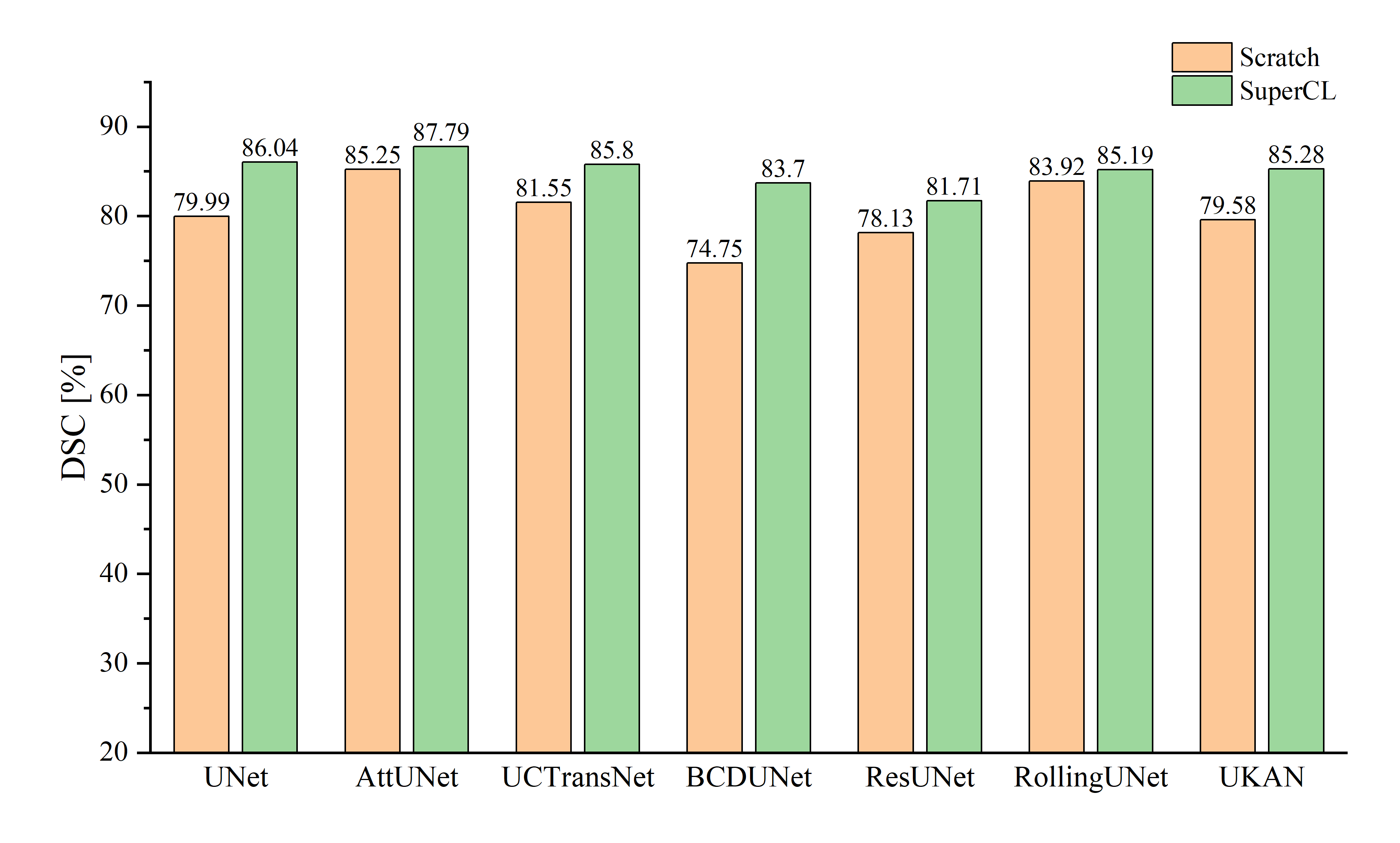}
  \caption{Comparison results of different variants of UNet backbone on ACDC with 25\% annotations.}
  \label{backbone}
\end{figure}

\noindent{\textbf{Variant U-Net backbones:}} The decoders used for medical image segmentation are usually U-shape, and our SuperCL can also be generalized to other variants of U-Net. We have conducted experiments on different variants of U-Net to verify generalization and universality of our proposed SuperCL. Specifically, we choose  AttUNet \cite{oktay2018attention} (decoder with attention blocks), UCTransNet \cite{wang2022uctransnet} (Transformer complement to CNN-based U-Net), BCDUNet \cite{azad2019bi} (Bi-directional ConvLSTM U-Net with densely connected convolutions), ResUNet \cite{drozdzal2016importance} (U-Net with residual connections),  
RollingUNet \cite{Liu_Zhu_Liu_Yu_Chen_Gao_2024} (incorporate MLP into U-Net) and UKAN \cite{li2024ukanmakesstrongbackbone} (integrate Kolmogorov-Arnold Networks into U-Net)
for comparison with the basic U-Net \cite{ronneberger2015u}. From Figure \ref{backbone}, we can see that our SuperCL can be generalized to different model backbones. Specifically, AttUNet even outperforms basic UNet. 

\begin{figure*}[h]
  \centering
  \includegraphics[width=\textwidth]{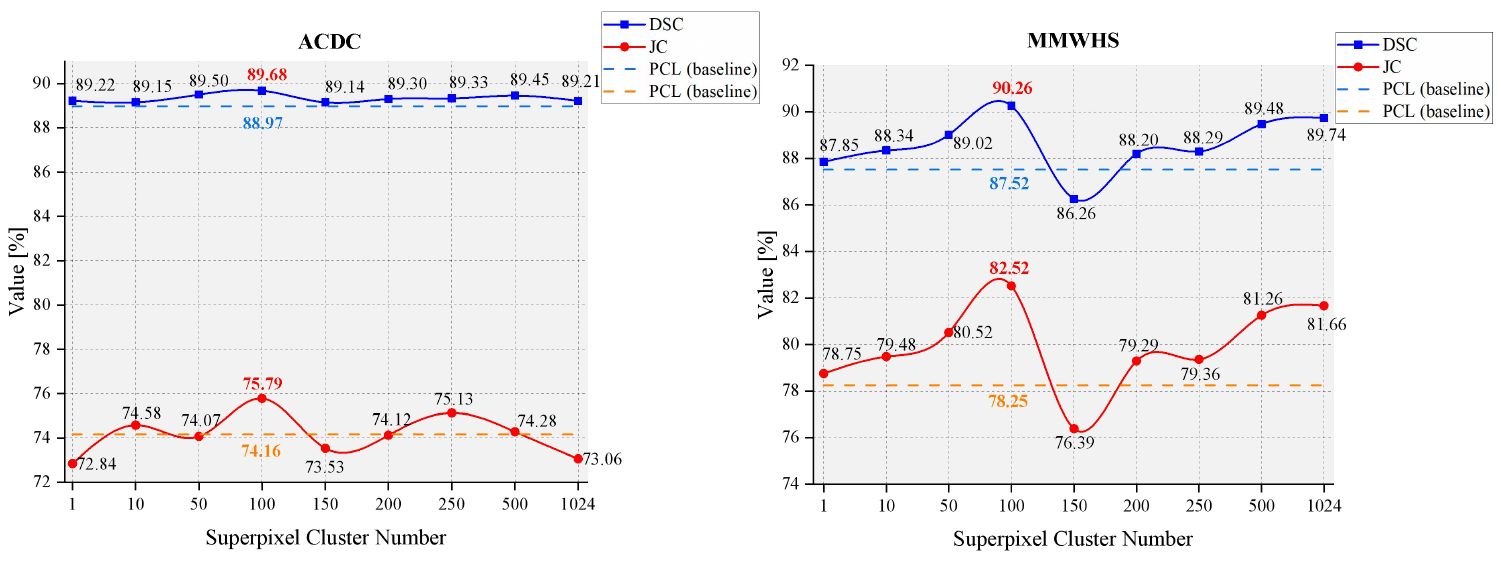}
  \caption{Comparison results of different superpixel cluster numbers on ACDC and MMWHS with 25\% annotations.}
  \label{sp_number}
\end{figure*}

\noindent{\textbf{Different numbers of contrastive pairs of ILCP:}}
When we apply ILCP to calculate $\mathcal{L}_{intra}$, there are totally 
$h \times w$ ($32 \times 32 = 1024$ in our setting) pixels as samples used for CL. And we resample the pixel-level
contrastive pairs using different strides to assess the best stride for SuperCL. According to the Table \ref{stride}, when stride = 1,
our proposed SuperCL achieves the best performance which is also used for comparison and ablation studies in the paper. However, if GPU memory is insufficient, you can also appropriately increase the stride to reduce the number of contrastive pairs, yet at the expense of segmentation performance. 

\noindent{\textbf{SuperCL can also serve as a good initialization in semi-supervised learning scenario:}} As we have declared in Introduction: ``Self-supervised learning offers a pre-training strategy that relies solely on unlabeled data to obtain a suitable initialization for training downstream tasks with limited annotations." And we have already conducted substantial experiments in self-supervised learning scenario. Furthermore, we want to testify the impact of our SuperCL in the semi-supervised learning scenario. Specifically, we utilize SuperCL to pre-train UNet encoder as the initialization of BCP \cite{bai2023bidirectionalcopypastesemisupervisedmedical} (CVPR 2023) and CauSSL \cite{10376893} (ICCV 2023) to find whether it can boost their segmentation performance in the semi-supervised scenario. We directly reported the results from CauSSL, which has made abundant comparison results with fully supervised learning (SL), and other 8 semi-supervised learning methods, including MT \cite{DBLP:journals/corr/TarvainenV17}, UA-MT \cite{DBLP:journals/corr/abs-1907-07034}, SASSNet \cite{DBLP:journals/corr/abs-2007-10732}, DTC \cite{DBLP:journals/corr/abs-2009-04448}, URPC \cite{DBLP:journals/corr/abs-2012-07042}, CPS \cite{DBLP:journals/corr/abs-2106-01226}, MC-Net+ \cite{DBLP:journals/corr/abs-2109-09960} and BCP \cite{bai2023bidirectionalcopypastesemisupervisedmedical} on ACDC dataset with 10\% / 20\% annotations. We follow the same training setting, and just utilize our SuperCL strategy to pre-train the UNet encoder with unlabeled CHD dataset as the initialization for both BCP and CauSSL. Then BCP and CauSSL use the semi-supervised training pipeline in their own paper, respectively to get segmentation performance on ACDC. As shown in Table \ref{SemiSL}, our SuperCL can really serve as a good initialization even in semi-supervised scenario, which can boost the segmentation performance of BCP and CauSSL on ACDC with 10\% / 20\% annotations. {\itshape i.e.} Compared with BCP, our SuperCL can get an improvement with 1.23\% DSC, 1.29\% JC, 2.21 HD95, 0.47 ASD on ACDC with 10\% annotations and an improvement with 0.41\% DSC, 0.69\% JC, 0.35 HD95, 0.11 ASD on ACDC with 20\% annotations. Compared with CauSSL, our SuperCL can boost it with an increase with 0.08\% DSC, 0.11\% JC, 1.58 HD95 0.18 ASD on ACDC with 10\% annotations and 0.36\% DSC, 0.55\% JC, 1.96 HD95, 0.38 ASD on ACDC with 20\% annotations. Meanwhile, with the assistance of our SuperCL, CauSSL with 20\% annotations is closer to SL (upper bound) with 100\% annotations.

\noindent{\textbf{Different numbers of superpixel cluster:}}
We make a gradient experiment across different numbers of superpixel cluster on ACDC and MMWHS with 25\% annotations, including two extreme situations: 1 single superpixel cluster and pixel-to-pixel 1024 (size of the feature map is $32\times32$) superpixel clusters. According to Fig. \ref{sp_number}, our SuperCL among all the cluster numbers except for 150 outpeforms the PCL baseline. And 100 clusters gets the best performance which is also used as our setting.

\begin{figure*}[t]
  \centering

  \subfloat[]{\includegraphics[width=\textwidth]{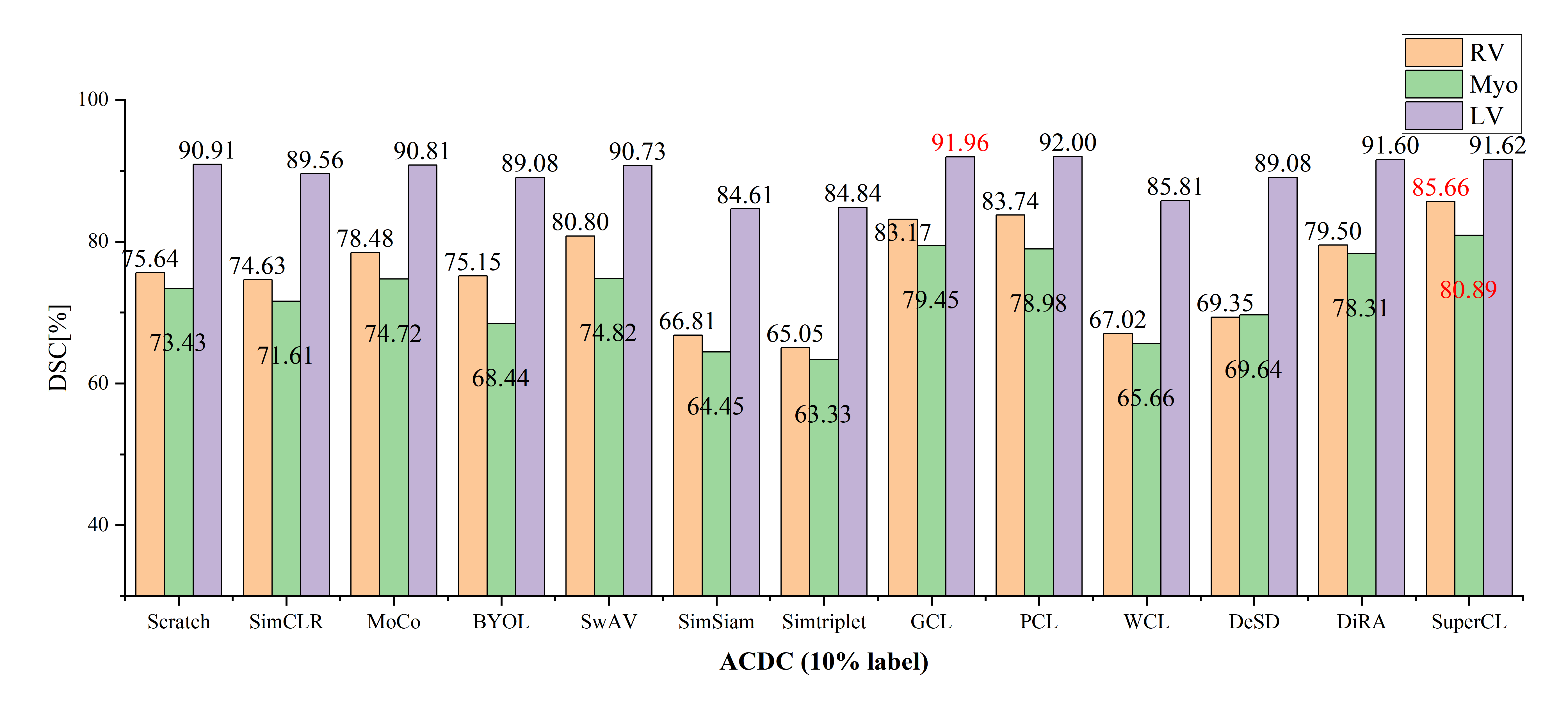}
  \label{fig:image1}}

  \vspace{1em} 

  \subfloat[]{\includegraphics[width=\textwidth]{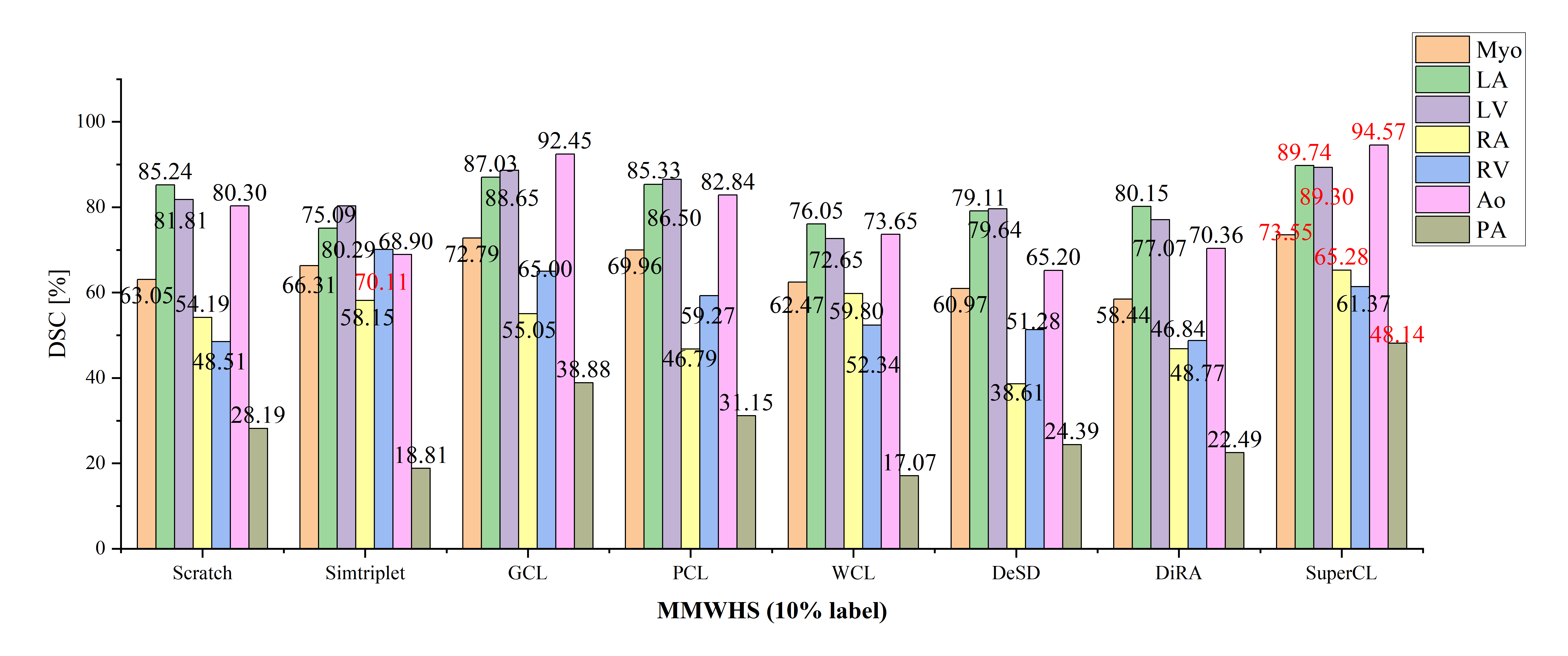}
  \label{fig:image2}}

  \caption{The performance (DSC) of each individual organ on ACDC and MMWHS with 10\% annotations.}
  \label{individual organ_1}
\end{figure*}

\begin{figure*}[t]
  \centering

  \subfloat[]{\includegraphics[width=\textwidth]{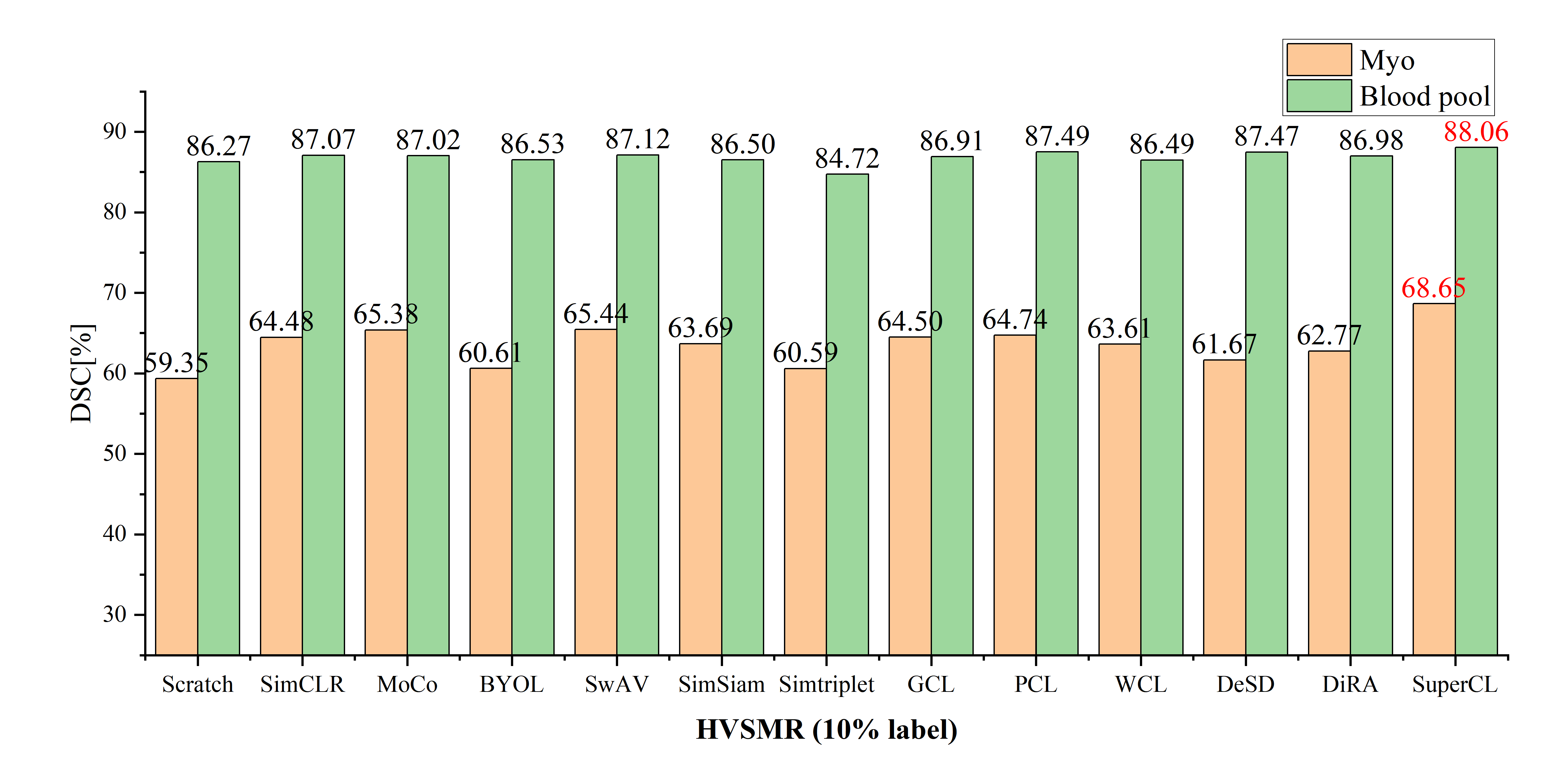}
  \label{fig:image1}}

  \vspace{1em} 

  \subfloat[]{\includegraphics[width=\textwidth]{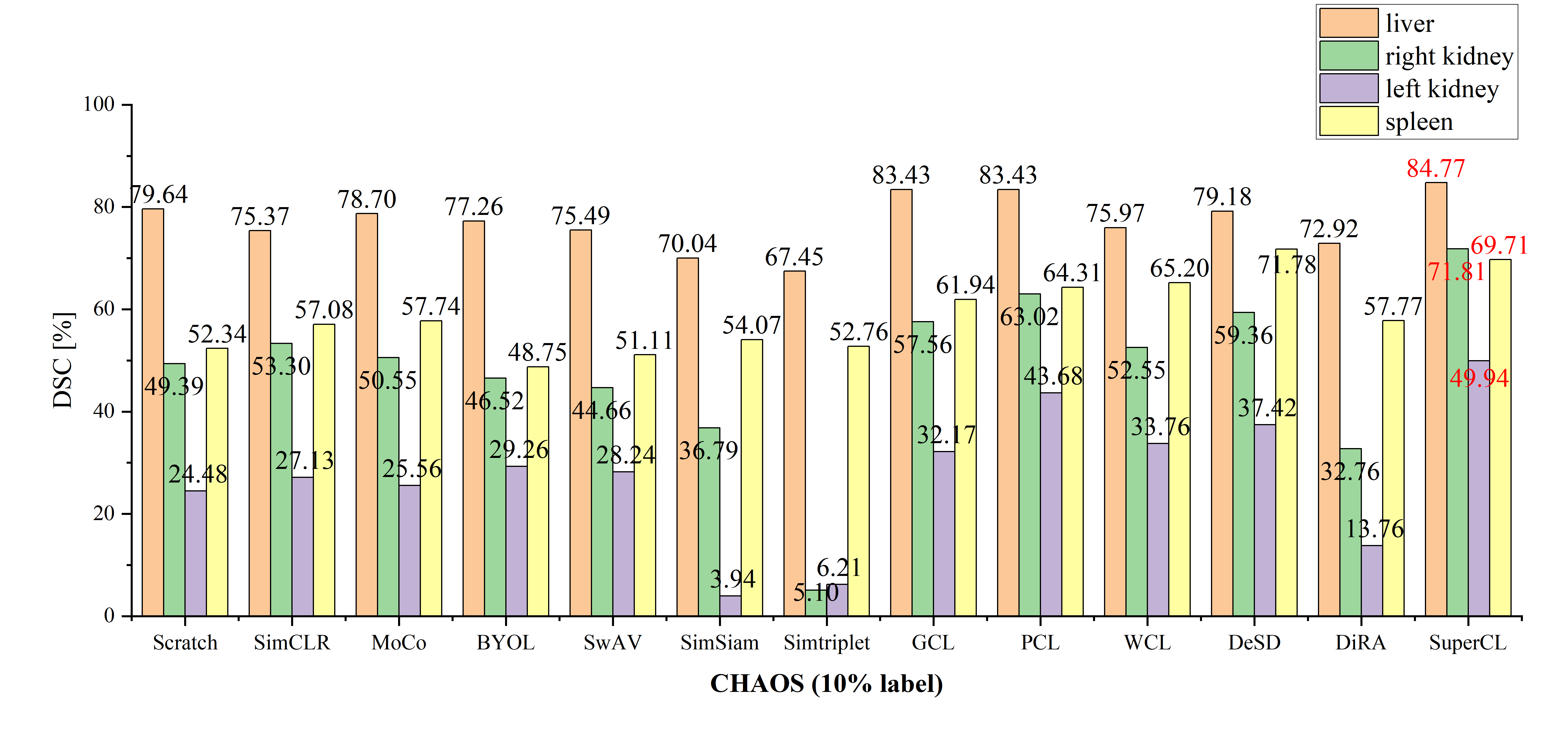}
  \label{fig:image2}}

  \caption{The performance (DSC) of each individual organ on HVSMR and CHAOS with 10\% annotations.}
  \label{individual organ_2}
\end{figure*}

\noindent{\textbf{Single Organ Segmentation Performance:}} 
We want further to testify the generalization ability of our proposed SuperCL on more detailed single organ segmentation of multi-organ datasets. As shown in Fig. \ref{individual organ_1} and \ref{individual organ_2}, our proposed SuperCL almost achieves all SOTA results (DSC) except for LV of ACDC and RV of MMWHS for individual organ segmentation on ACDC, MMWHS, HVSMR and CHAOS with 10\% annotations. And our SuperCL can gain a huge improvement compared with other baseline methods, {\itshape e.g.} RV: SuperCL 85.66\% vs. PCL 83.74\% (1.92\% $\uparrow$) on ACDC; LA: SuperCL 89.74\% vs. GCL 87.03\% (2.71\% $\uparrow$), Ao: SuperCL 94.57\% vs. GCL 92.45\% (2.12\% $\uparrow$) on MMWHS; Myo: SuperCL 68.65\% vs. SwAV 65.44\% (3.21\% $\uparrow$) on HVSMR and liver: SuperCL 84.77\% vs. PCL 83.43\% (1.34\% $\uparrow$) on CHAOS.

\section{Conclusion}

In this work, we propose a novel CL approach named SuperCL for medical image segmentation pre-training which exploits the structural prior and pixel correlation of images by introducing two contrastive pairs generation strategies: ILCP and IGCP, along with two novel modules: ASP and CCL. Experiments on multiple medical image datasets indicate our SuperCL outperforms existing SOTA CL strategies.

\bibliographystyle{IEEEtran}
\bibliography{TNNLS.bib}
\section{Appendix}
\subsection{Network Architecture}

Our proposed SuperCL framework has two parts: an encoder $e(\cdot)$ and a projector $g(\cdot)$. And during downstream finetuning, a decoder $d(\cdot)$ will be used to form the whole segmentation network. The encoder and decoder are based on a UNet architecture and the projector is an image-level instance projector.
Specifically, the encoder $e(\cdot)$ consists of 4 convolutional blocks and a center block. Each convolutional block consists of two "$\textbf{Conv2d} \rightarrow \textbf{BatchNorm2d} \rightarrow \textbf{LeakyReLU}$" structures followed by a $2\times 2$ \textbf{Maxpooling} layer with stride 2. The kernel size of Conv2d is $3\times 3$ with 1 zero padding. 
Similar to the encoder, the decoder of UNet also consists of 4  convolutional blocks and a projection head. Each convolutional block consists of a \textbf{ConvTranspose2d} followed by two "$\textbf{Conv2d} \rightarrow \textbf{BatchNorm2d} \rightarrow \textbf{LeakyReLU}$" structures which is the same as those in the encoder. The projection head is a simple \textbf{Conv2d} structure which will project the fused features into a segmentation map. 
The image-level instance projector consists of an "$\textbf{AdaptiveAvgPool2d} \rightarrow \textbf{Flatten} \rightarrow \textbf{Linear} \rightarrow \textbf{ReLU} \rightarrow \textbf{Linear}$" structure. The kernel size of Conv2d is still $3\times 3$ with 1 zero padding.

B (batch size) = 16, C (Channels) = 128, h (height of the feature map) = 32, w (width of the feature map) = 32.



\begin{figure*}[h]
  \centering
  \includegraphics[width=0.8\textwidth]{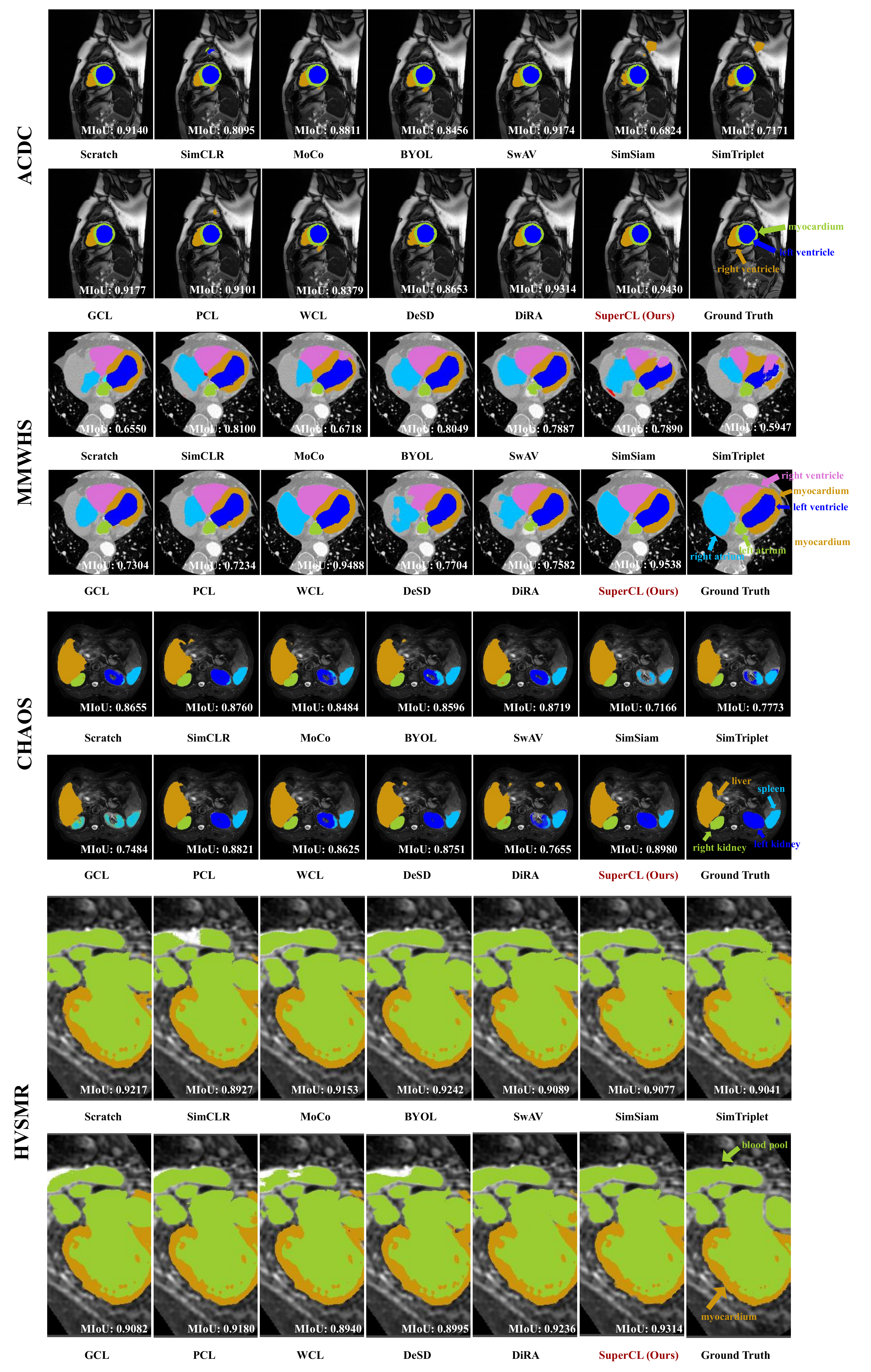}
  \caption{Visualization of more multi-organ segmentation results on ACDC, MMWHS, CHAOS and HVSMR.}
  \label{multi_organ_1}
\end{figure*}

\begin{figure*}[h]
  \centering
  \includegraphics[width=0.8\textwidth]{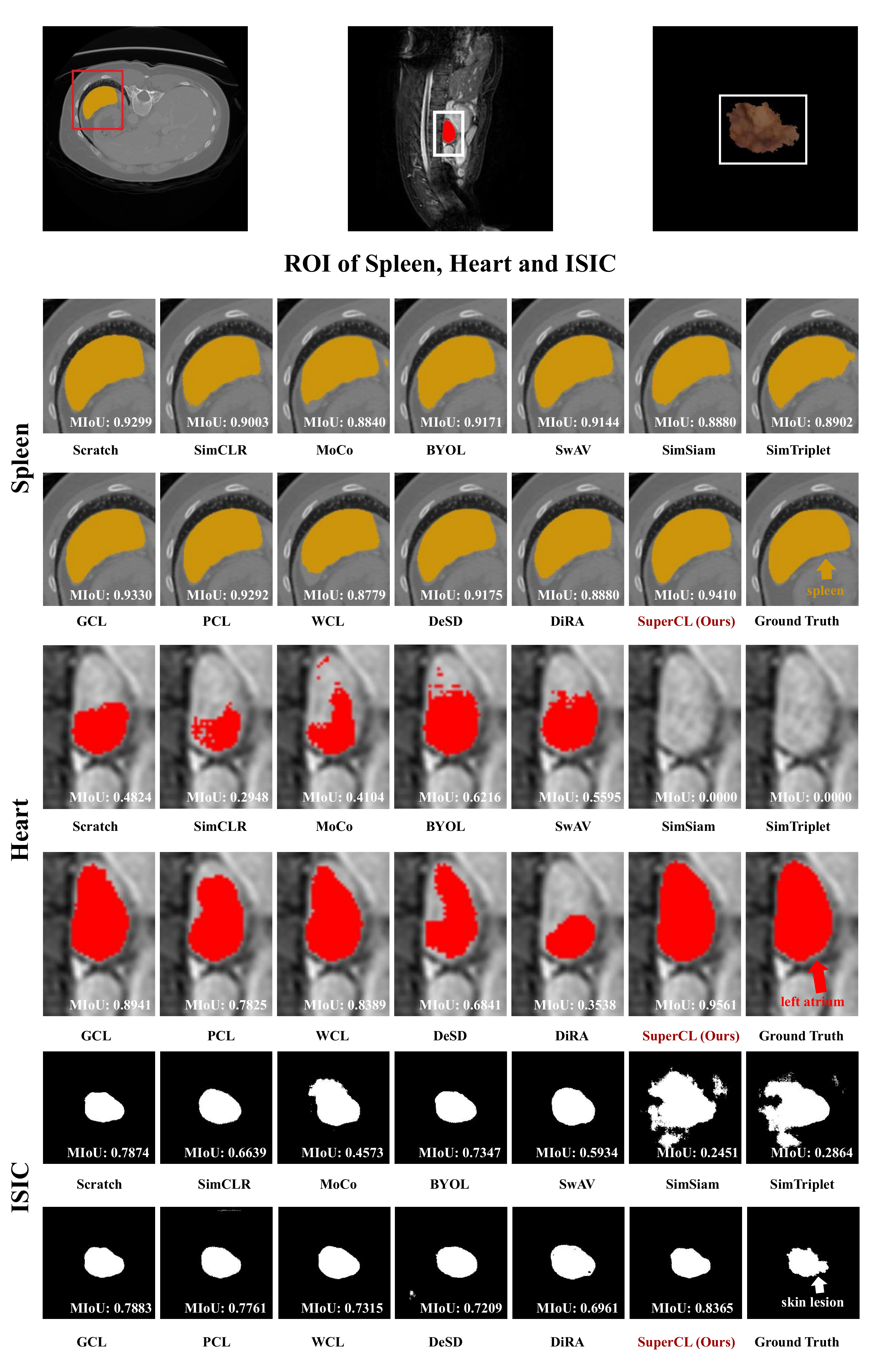}
  \caption{Visualization of ROI-based segmentation results on Spleen, Heart and Spleen.}
  \label{ROI_1}
\end{figure*}
\end{document}